\documentclass[conference]{IEEEtran}
\IEEEoverridecommandlockouts

\IEEEoverridecommandlockouts 
\usepackage[dvips]{graphicx}
\usepackage{algorithmic}
\usepackage{algorithm}
\usepackage[frozencache,cachedir=minted-cache]{minted}
\usepackage{amsmath,amsthm}
\usepackage{url}
\usepackage{multirow}

\usepackage{amsfonts}
\usepackage{bbm}
\usepackage{subcaption}
\newtheorem{definition}{Definition}

\newtheorem{example}{Example}

\newcommand{\nn}{f}
\newcommand{\nnL}{f^L}
\newcommand{\nnU}{f^U}
\newcommand{\volume}{\textnormal{vol}}

\newcommand{\surrogate}{g}
\newcommand{\Surrogate}{G}
\newcommand{\inset}{\mathbb{R}^n}
\newcommand{\outset}{\mathbb{R}^m}
\newcommand{\dataset}{\mathcal{D}}

\newcommand{\inpoint}{x}
\newcommand{\outpoint}{y}

\newcommand{\advpert}{\mathbb{B}_p(x,\epsilon)}
\newcommand{\pert}{\epsilon}
\newcommand{\limedist}{d_x}
\newcommand{\precond}{\phi_{\mathsf{pre}}}
\newcommand{\postcond}{\phi_{\mathsf{post}}}

\newcommand{\linmat}{W^{(i)}}
\newcommand{\linbias}{b^{(i)}}
\newcommand{\vechat}{\hat{z}}
\newcommand{\vecact}{z}
\newcommand{\hiddvec}{\hat{z}^{(i)}}
\newcommand{\hiddvecact}{z^{(i)}}
\newcommand{\hiddvecinput}{z^{(i-1)}}
\newcommand{\inputvec}{z^{(0)}}

\newcommand{\concretelowersingle}{l}
\newcommand{\concreteuppersingle}{u}
\newcommand{\preact}{h}
\newcommand{\postact}{a}
\newcommand{\nnslopesingle}{\alpha}
\newcommand{\activar}{\delta}

\DeclareMathOperator*{\argmax}{arg\,max}
\DeclareMathOperator*{\argmin}{arg\,min}

\title{When to Trust AI: Advances and Challenges for Certification of Neural Networks}
\author{
\IEEEauthorblockN{Marta Kwiatkowska, Xiyue Zhang}
\IEEEauthorblockA{0000-0001-9022-7599\\
0000-0003-1649-7165\\
University of Oxford, UK\\
Email: \{marta.kwiatkowska, xiyue.zhang\}@cs.ox.ac.uk}
}

\begin{document}
\maketitle              

\begin{abstract}
Artificial intelligence (AI) has been advancing at a fast pace and it is now poised for deployment in a wide range of applications, such as autonomous systems, medical diagnosis and natural language processing. Early adoption of AI technology for real-world applications has not been without problems, particularly for neural networks, which may be unstable and susceptible to adversarial examples. In the longer term, appropriate safety assurance techniques need to be developed to reduce potential harm due to avoidable system failures and ensure trustworthiness. Focusing on certification and explainability, this paper provides an overview of techniques that have been developed to ensure safety of AI decisions and discusses future challenges.
\end{abstract}

\section{Introduction}

\begin{figure}[!t]
     \centering
     \begin{subfigure}[b]{0.25\linewidth}
         \centering
    \includegraphics[height=2.5cm]{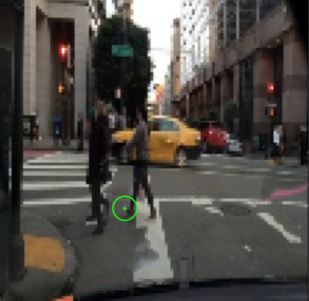}
         \label{fig:intro_1}
     \end{subfigure}
     \quad
     \begin{subfigure}[b]{0.25\linewidth}
         \centering
         \includegraphics[height=2.5cm]{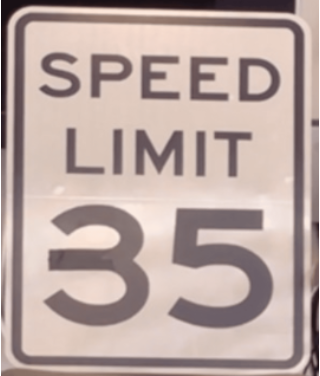}
         \label{fig:intro_2}
     \end{subfigure}
     \begin{subfigure}[b]{0.25\linewidth}
         \centering
         \includegraphics[height=2.5cm]{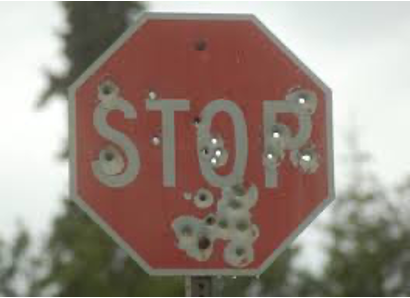}
         \label{fig:intro_3}
     \end{subfigure}
        \caption{Challenges of safe traffic sign recognition. Single-pixel adversarial attack from \cite{WHK2017} (left), physical attack (middle) and a real traffic sign (right).}
\label{fig:intro}
\end{figure}

Artificial intelligence (AI) has advanced significantly in recent years, largely due to the step improvement enabled by deep learning in data-rich tasks such as computer vision or natural language processing. AI technologies are being widely deployed and enthusiastically embraced by the public, as is evident from the take up of ChatGPT and Tesla. However, deep learning lacks robustness, and neural networks (NNs), in particular, are unstable with respect to so called {\em adversarial perturbations}, often imperceptible modifications to inputs that can drastically change the network’s decision. Many such examples have been reported in the literature and the media. Figure~\ref{fig:intro} (left) shows a dashboard camera image from \cite{WHK2017}, for which a change of a single pixel to green changes the classification of the image from red traffic light to green, which is potentially unsafe if there is no fallback safety measure; while this is arguably an artificial example, some modern cars have been observed to mis-read traffic signs, including the physical attack in Figure~\ref{fig:intro} (middle), where the digit 3 has been modified. Traffic sign recognition is a complex problem to specify and solve, see Figure~\ref{fig:intro} (right), which shows a real traffic sign in Alaska. As with any maturing technology, it is natural to ask if AI is ready for wide deployment, and what steps – scientific, methodological, regulatory, or societal – can be taken to achieve its trustworthiness and reduce potential for harm through rushed roll-out. This is particularly important given the fast-paced development of AI technologies and the natural propensity of humans to overtrust automation. 

For AI to be trusted, particularly in high-stakes situations, where avoidable failure or wrong decision can lead to harm or high cost being incurred, it is essential to provide {\em provable guarantees} on the critical decisions taken autonomously by the system. Traditionally, for software systems this has been achieved with {\em formal verification} techniques, which aim to formally prove whether the system satisfies a given specification, and if not provide a diagnostic counter-example. Founded on logic, automated verification, also known as model checking, achieves this goal by means of executing a verification algorithm on a suitably encoded model of the system. Software verification has become an established methodology and a variety of tools of industrial relevance are employed in application domains such as distributed computation, security protocols or hardware. Beginning with~\cite{DLV,Katz17}, over the past few years a number of formal verification techniques have been adapted to neural networks, which are fully data-driven and significantly differ from the state-based transition system models of conventional software, and have given rise to practical, algorithmic techniques that provide provable guarantees on neural network decisions~\cite{VNNComp}.

This paper aims to provide an overview of existing techniques that can be used to increase trust in AI systems and outline future scientific challenges, while at the same time raising awareness of potential risks with early adoption. It is taken as granted that safety assurance of AI systems is complex and needs to involve appropriately regulated processes and assignment of accountability. The topics discussed in this paper are by no means exhaustive, but offer a representative selection of techniques and tools that can be used within such safety assurances processes, and can be adapted, extended or built upon to increase robustness and trustworthiness of AI systems. The paper will focus on highlighting the following two aspects:

\begin{itemize}
\item {\bf Certification}: focusing on individual decisions (possibly critical to the integrity of the system) that are made by neural networks, we provide an overview of the main methodological approaches and techniques that have been developed to obtain provable guarantees on the correctness of the decision, which can thus be used for certification. The sources of computational complexity of neural network verification will be discussed, as well as limitations of existing methods and ways to address them.

\item {\bf Explainability}: neural networks are ‘black boxes’ that are trained from data using obscure optimization processes and objectives, and it is argued that users of AI systems will benefit from the ability to obtain explanations for the decisions. We summarise the main approaches to producing explanations and discuss that they may lack robustness and how this issue can be addressed.
\end{itemize}

The overview includes high-level description of main algorithms, which are illustrated by worked examples to explain their behaviour to the interested reader.
This is followed by a selection of case studies of robustness analysis and/or certification drawn from a variety of
application domains, with the aim to highlight the strengths and weaknesses of the approaches. Finally, future challenges and suggestions for fruitful directions to guide the developments in this actively studied and important area will be outlined.

The paper is organised as follows. Section \ref{sect-II} introduces the main concepts, focusing on neural networks in the supervised learning setting. Section \ref{sect-III} provides an overview of the main (forward and backward) analysis approaches, with a description of the working for a selection of algorithms illustrated by worked examples. Section \ref{sect-IV} includes a few excerpts from a selection of verification and certification experiments, aimed at highlighting the uses of the main methods,  and Section \ref{sect-V} outlines future challenges. Finally, Section \ref{sect-VI} concludes the paper.

\section{Safety, Robustness and Explainability}\label{sect-II}

In the context of safety-critical systems, safety assurance techniques aim to prevent, or minimise the probability of, a hazard occurring, and appropriate safety measures are invoked in case of failures. In this paper, we focus on critical decisions made by neural networks, which we informally refer to as safe if they satisfy a given property, which can be shown or disproved by formal verification.  
Before discussing formal verification techniques, we 
begin with background introduction to the main concepts of deterministic neural networks, 
their (local) robustness and explanations. 

\subsection{Neural Networks}

We consider neural networks in the supervised learning setting. 
A neural network is a function $\nn: \inset \rightarrow \outset$ mapping from the input space to the output space,  which is typically trained based on a dataset $\dataset$ of pairs $(\inpoint,\outpoint)$ of input  $\inpoint$ and ground truth label $\outpoint$. 
A neural network consisting of $\mathit{L}+1$ layers (including the input layer) can be characterized by a set of matrices $\{\linmat\}_{i=1}^{L}$ and bias vectors $\{\linbias\}_{i=1}^{L}$ for linear (affine) transformations, followed by pointwise activation functions, such as $\mathit{ReLU}$, $\mathit{Sigmoid}$, and $\mathit{Tanh}$, for  nonlinear transformations. 
We use $\hiddvec$ and $\hiddvecact$ to denote the pre-activated and activated vectors of the $i$-th layer, respectively.
The layer-by-layer forward computation of neural networks can be described as follows:
\begin{itemize}
    \item \textit{Linear transformation.} The linear transformation generates a pre-activated vector $\hiddvec = \linmat\cdot \hiddvecinput+\linbias$ ($i \in  [1,L]$) from the output of the previous layer, and $\inputvec=x$ denotes the input vector.
    \item \textit{Pointwise nonlinear transformation.} The pointwise nonlinear transformation generates the activation vector $\hiddvecact = \sigma(\hiddvec)$ ($i \in  [1,L]$). In practice, $\mathit{softmax}$ is usually employed as the activation function for the output layer in classification tasks, which provides the normalised relative probabilities of classifying the input into each label.
\end{itemize}
Given an input $x \in \inset$, the output of $f$ on $x$ is defined by $f(x)= f^{(L)} \circ \cdots \circ f^{(1)}(x)$, where $f^{(i)}$ denotes the mapping function of the $i$-th layer, which is the composition of linear and pointwise nonlinear transformations. 

\begin{figure}[tbp]
\centering
\includegraphics[width=0.3\textwidth]
     {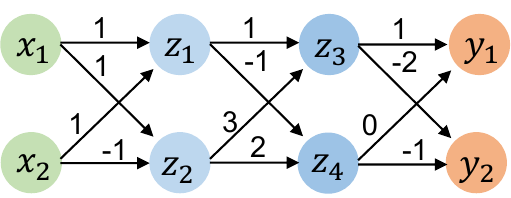}
\caption{A feed-forward neural network.}
\label{fig:net_demo}
\end{figure}

\begin{example}\label{eg:net_demo}
Figure \ref{fig:net_demo} shows a simple feed-forward (and fully connected (FC)) neural network with four layers and ReLU as the activation function.
$x_{1}$,  $x_{2}$ represent two input neurons. $z_{1}$, $z_{2}$ and $z_{3}$, $z_{4}$ represent the activated neurons of the two hidden layers. $y_{1}$,  $y_{2}$ are two output neurons.
The forward computation from the input layer to the output layer is as follows (ReLU is denoted as $\sigma$). 
\begin{align}
     & \vecact_{1}=\sigma(x_{1}+x_{2}), \quad \vecact_{2}=\sigma(x_{1}-x_{2}) \\
     & \vecact_{3} = \sigma(\vecact_{1} + 3\vecact_{2}), \quad 
     \vecact_{4} = \sigma(-\vecact_{1} +2 \vecact_{2})\\
     & y_1 = \vecact_{3}, \quad
     \quad y_2 = -2\vecact_{3} - \vecact_{4}
\end{align}
We will use this neural network as a running example to illustrate different problem formulations and methods to address them.
\end{example}

\begin{figure*}[tbp]
    \centering
    \includegraphics[width=1\textwidth]{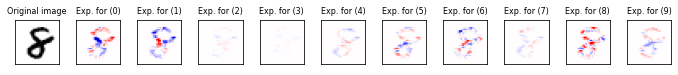}
    \caption{The IG explanation for each of the classes of the MNIST dataset, where red indicates a positive contribution and blue a negative. Figure taken from \cite{Falconmore22}. }
    \label{fig:ig_mnist_exp}
\end{figure*}

\subsection{Robustness}
Robustness focuses on neural networks' resilience to adversarial attacks, noisy input data, etc., at test time, known as evasion attacks \cite{Biggio13,Szegedy13,Biggio18}. 
Attacks at training time are known as poisoning attacks \cite{Biggio12Poison}, which have been omitted from this overview.

{\em Adversarial robustness} \cite{Szegedy13} of neural networks formalizes the desirable property that a well-trained model makes consistent predictions when its input data point is subjected to small adversarial perturbations. 
{\em Local (adversarial) robustness} pertains to a given input point $\inpoint$ with ground truth label $\outpoint$, and is usually defined in terms of invariance of the network’s decision within a small neighbourhood $\advpert$ of $\inpoint$, for a class of perturbations bounded by $\pert$ with respect to the $\ell_p$ norm. 
\begin{definition}
Given a (deterministic) neural network $\nn$, a labelled input data point $(\inpoint,\outpoint)$, and a perturbation bound $\pert$, the local robustness property of $\nn$ on $\inpoint$ is defined as 
$$\forall x' \in \advpert. \,\argmax_{i=1,\cdots,m} f_i(x')=y,$$ where $\advpert$ denotes the adversarial $\ell_p$-ball of radius $\epsilon$ around input $x$. 
\end{definition}
Should there exist a point $x'$ in the neighbourhood whose class is different than $y$, it is referred to as an {\em adversarial example}. 

A related concept is that of a {\em maximal safe radius (MSR)}~\cite{Wu2020}, denoted $MSR(x)$, which is the minimum distance from $x \in \inset$ to the decision boundary, and is defined as the largest $\epsilon >0$ such that 
$\forall x' \in \advpert. \,\argmax_{i=1,\cdots,m} f_i(x')= \argmax_{i=1,\cdots,m} f_i(x)$. 
Computing the value of MSR, say $\gamma$, provides a {\em guarantee} that the decision is robust (safe) for perturbations up to $\gamma$. On the other hand, finding an adversarial example at distance $\gamma'$ is witness to the failure of robustness.

Global robustness~\cite{Leino21globalrobust} concerns the stability of predictions over the whole input space and is omitted.

\subsection{Explainability}

Explainability \cite{Du20Interpretable,molnar2020interpretable} aims to understand and interpret why a given neural network makes certain predictions.
The term explainability is often used interchangeably with interpretability in the literature, though interpretability usually refers to explaining how the model works.
In this overview, we focus on local (pointwise) explainability for an individual \emph{model decision}, which is categorised into  {\em feature attribution} methods, which heuristically estimate feature attribution scores for model predictions and include \textit{gradient-based} \cite{bach2015pixel, sundararajan2017axiomatic} and \textit{perturbation-based} techniques \cite{ribeiro2016should,ribeiro2018anchors}, 
and {\em abduction-based} methods~\cite{IgnatievNM19,LZM+21}, which identify the features that imply the decision and can thus provide (safety) guarantees.
Attribution scores can also be used for feature importance ranking \cite{Janusz23attribute} to provide an overall understanding of the importance of different input attributes on the model decisions. 
\subsubsection{Gradient-based methods} 
Gradient-based methods aim to 
estimate feature attribution scores for model predictions.
Among these, a prominent method is the integrated gradients (IG)~\cite{sundararajan2017axiomatic}, 
which measures the attribution score of each input feature to the model's prediction by integrating the gradients of the model's output with respect to the input features along the path from a baseline input to the actual input. 

\begin{definition}
    Given a neural network $\nn$, an input $x$ and a baseline input $x'$, the integrated gradients for each input feature $i \in [1,\cdots,n]$ are defined as the weighted (by input feature difference) integral of the gradients over the straight line path between $x$ and $x'$:
\begin{equation}
    \label{eq:ig}
\text{IG}_i(x) = (x_i - x'_i) \times \int_{\alpha=0}^{1} \frac{\partial f(x'+\alpha \times (x-x'))}{\partial x_i}d\alpha
\end{equation}
\end{definition}
Figure~\ref{fig:ig_mnist_exp} from \cite{Falconmore22} presents an illustrative example of IG explanations, showing the explanation for each class of a correctly classified handwritten-digit ``8'' from the MNIST dataset. In this example, positive contributions are highlighted in red, while negative contributions are indicated by blue.

\subsubsection{Perturbation-based methods}
LIME \cite{ribeiro2016should} and its successor Anchors \cite{ribeiro2018anchors} are representatives of explainability methods that deploy a perturbation-based strategy to generate local explanations for model predictions.
LIME assumes local linearity
in a small area around an input instance and generates a set of synthetic data by perturbing the original input. 
Anchors \cite{ribeiro2018anchors} explains the model predictions by identifying a set of decision rules that ``anchors'' the prediction. Compared with LIME, Anchors generates more explicit decision rules and derives local explanations by consulting $x$'s perturbation neighbourhood in different ways. 
In particular, 
Anchors evaluates the coverage fraction of the perturbed data samples sharing the same class as $x$, matching the decision rules.

\subsubsection{Robust explanations}
The explanation techniques mentioned above use different heuristics to derive local explanations, demonstrating effective generality beyond the given input but lacking robustness to adversarial perturbations. 
The robustness notion for explanation is important to ensure the stability of the explanation in the sense that the explanation is logically sufficient to imply the prediction.
Intuitively, the computed explanation for a perturbed input should remain the same as the original input.

To this end,
\cite{IgnatievNM19,LZM+21} introduce a principled approach to derive explanations with formal guarantees by 
exploiting abduction reasoning.
This ensures the robustness of the explanation by requiring its invariance w.r.t. any perturbation of the remaining features that are left out.
The explanation method of \cite{LZM+21} focuses on \textit{optimal robust explanations (OREs)}, to provide both robustness guarantees and optimality w.r.t. a cost function.
Optimality provides the flexibility to control the desired properties of an explanation. For instance, the cost function could be defined as the length of the explanation to derive minimal but sufficient explanations.

\section{Certification for Neural Networks}\label{sect-III}
In this section, we present an overview of recent advances for certification of neural networks, with a focus on formal verification.
Given a neural network $\nn: \inset \rightarrow \outset$, we consider the formal verification problem \cite{VNNComp}, defined for a property specified as a pair $(\precond, \postcond)$ of 
precondition and postcondition, by requiring that 
$\forall x \in \inset. \,x \models \precond \implies f(x) \models \postcond$, that is, for all inputs satisfying the precondition the corresponding (optimal softmax) decision must satisfy the postcondition. 
Typically, $\precond \subseteq \inset$ and $\postcond \subseteq \outset$, but can be respectively induced from subsets of input features or sets of labels. 
Formal verification then aims to establish algorithmically whether this property holds, thus resulting in a {\em provable guarantee}.  Otherwise, the property may be falsified, in which case a witness is provided, or inconclusive.
Sometimes, we may wish to compute the proportion of inputs that satisfy the postcondition, known as {\em quantitative verification} \cite{wang2021statistically}.

Various formal verification methods have been proposed to provide provable guarantees for neural networks. 
We classify existing verification methods into forward and backward analysis, depending on whether they start from the input or output space.

\begin{itemize}
    \item \textit{Forward analysis:} Forward analysis methods start from the precondition $X = \{x \in \inset ~|~ x \models \precond\}$ defined on the input space, and check whether the outputs (corresponding to the input region) satisfy the postconditions $\postcond$. For example, robustness verification approaches \cite{Gehr18, Wang18, Singh19, Wang21beta} start from the perturbation neighbourhood of a given input, e.g., an $l_{\infty}$ ball around an input point $x$, and compute bounds on the outputs to check whether the predicted labels over the adversarial region are preserved. 
    \item \textit{Backward analysis:} Backward analysis methods start from the postcondition $Y = \{y \in \outset ~|~ y \models \postcond\}$ and aim to find the set of inputs that lead to such outputs. 
    For example, preimage generation (inverse abstraction) approaches \cite{Matoba20Exact, Dathathri19Inverse, Kotha23BoundPreimage, ZWK23}  start from the output constraints, e.g., a polytope constraining the probability of the target label is greater than the other labels, and derive the input set that provably leads to this particular decision.  
\end{itemize}
We remark that, similarly to formal verification for conventional software, certification for machine learning models is computationally expensive, and it is therefore recommended for use in safety- or security-critical settings. In less critical situations,  diagnostic methods  \cite{Janusz23Diagnosis}, which approximate model decisions to analyse their predictions, can be employed to investigate both model- and data-related issues.

\subsection{Forward Analysis Methods}

We categorize the forward analysis methods into two groups:
\textit{sound but incomplete} and \textit{complete} methods.
Soundness and completeness are essential properties of verification algorithms, which are defined as follows. 
\begin{itemize}
    \item   \textit{Soundness:} A verification algorithm is sound if the algorithm returns True and the verified property holds. 
    \item \textit{Completeness:} A verification algorithm is complete if (i) the algorithm never returns unknown; and (ii) if the algorithm returns False, the property is violated.
\end{itemize}

\begin{figure}[!t]
     \centering
     \begin{subfigure}[b]{0.25\linewidth}
         \centering
         \includegraphics[width=\textwidth]{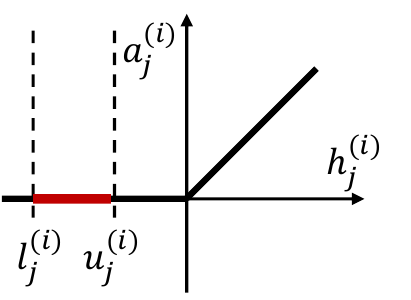}
         \label{fig:negative}
     \end{subfigure}
     \quad
     \begin{subfigure}[b]{0.25\linewidth}
         \centering
         \includegraphics[width=\textwidth]{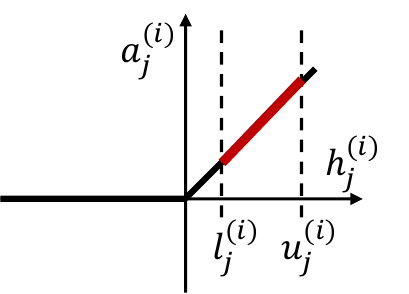}
         \label{fig:positive}
     \end{subfigure}
     \quad
     \begin{subfigure}[b]{0.25\linewidth}
         \centering
         \includegraphics[width=\textwidth]{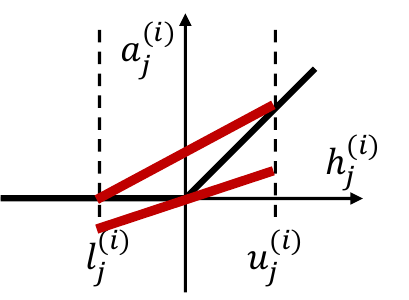}
         \label{fig:unstable}
     \end{subfigure}
        \caption{Illustration of the convex relaxation for inactive (left), active (middle) and unstable (right) ReLU neurons.}
\label{fig:linear_relaxation}
\end{figure}
\subsubsection{Incomplete methods}
Incomplete verification methods leverage approximation techniques, such as search \cite{WHK2017,Wu2020}, convex relaxation \cite{WongK18} and abstract interpretation \cite{CousotC92}, respectively to compute lower/upper bounds on MSR or the non-convex optimization problem. 
A safety property is verified when the reachable outputs satisfy the postcondition; otherwise, no conclusion can be drawn. 
At the same time, due to the relaxation introduced by the approximation techniques, incomplete methods have better scalability than complete ones.

\paragraph{Game-based search} 
Knowledge of the maximum safe radius (MSR) can serve as a guarantee on the maximum magnitude of the allowed adversarial perturbations. Unfortunately, MSR computation is intractable, and instead approximate algorithms have been developed for images in \cite{Wu2020}, and extended to videos in \cite{Wu2019}, that compute lower and upper bounds on MSR with provable guarantees, i.e., bounded error. The method relies on the network satisfying the Lipschitz condition and can be configured with a variety of feature extraction methods, for example SIFT. Given an over-approximation of the Lipschitz constant, the computation is reduced to a finite optimization over a discretisation of the input region $X$ corresponding to the precondition $\precond$. The resulting finite optimization is solved in anytime fashion through a two-player game, where player 1 selects features and player 2 perturbs the image representation of the feature, and the objective is set to minimise the distance to an adversarial example. 
Under the assumptions, the game can be unfolded into a finite tree and Monte Carlo Tree Search (MCTS) used to approximate MSR upper bound, and Admissible A* MSR lower bound, respectively.

\paragraph{Bound propagation}
A common technique for incomplete verification is applying convex relaxation to bound nonlinear constraints in neural networks.
This way, the original non-convex optimization problem is transformed into a linear programming problem.
With the relaxed linear constraints, the global lower and upper bounds can be computed more efficiently for the associated (relaxed) linear program. 
Representative methods that adopt efficient bound propagation include convex outer adversarial polytope \cite{WongK18}, CROWN \cite{zhang2018crown} and its generalization \cite{xu2021fast, Wang21beta}.
Figure \ref{fig:linear_relaxation} illustrates convex relaxation using linear bounding functions to bind ReLU neurons. 
Note that relaxation is only introduced for unstable neurons, while the ReLU constraints for inactive and active ones are exact.
For unstable neurons, the lower and upper bounding function for the $j$-th neuron of the $i$-th layer $\postact^{(i)}_j(\inpoint)$ (activated value) with regard to $\preact^{(i)}_j(\inpoint)$ (before activation) are: 
\begin{equation}\label{eq:bound_fun}
\nnslopesingle^{(i)}_j \preact^{(i)}_j(\inpoint)  \le \postact^{(i)}_j(\inpoint) \le -\frac{\concreteuppersingle^{(i)}_j\concretelowersingle^{(i)}_j}{\concreteuppersingle^{(i)}_j - \concretelowersingle^{(i)}_j}
    + \frac{\concreteuppersingle^{(i)}_j}{\concreteuppersingle^{(i )}_j - \concretelowersingle^{(i )}_j} \preact^{(i)}_j(\inpoint)
\end{equation}
where a flexible lower bound function with parameter $\nnslopesingle^{(i)}_j$ as in \cite{xu2021fast} is used, which leads to a valid lower bound for any parameter value within $[0,1]$. 

By propagating the linear (symbolic) upper and lower bounds layer by layer, we can obtain the linear bounding functions $\nnL$, $\nnU$ for 
the entire neural network $\nn$, and it holds that $\forall x \in X. \,\nnL(x) \le \nn(x) \le \nnU(x)$.
The non-convex verification problem is thus transformed into a linear program with the objective linear in the decision variables. 
The certified upper and lower bounds can be computed by taking the maximum, $\max_{x \in \advpert}\nnU(x)$, and the minimum,  $\min_{x \in \advpert}\nnL(x)$, which have \textit{closed-form} solutions for linear objectives ($\nnU$, $\nnL$) and convex norm constraints $\advpert$.

\begin{figure}[tbp]
\centering
\includegraphics[width=0.5\textwidth]
     {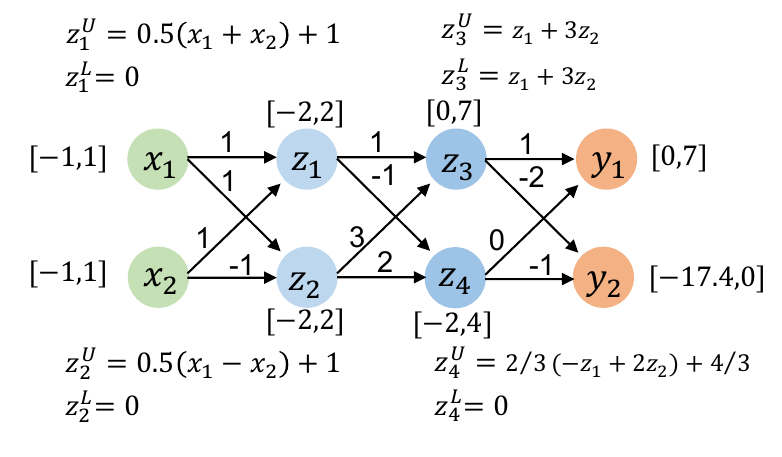}
\caption{Verification via bound propagation.}
\label{fig:demo_boundprop}
\end{figure}
\begin{example}\label{eg:boundProp}
Consider the neural network illustrated in Example \ref{eg:net_demo}.
The verification problem we consider is 
given by the pre-condition $\precond =\{x \in \mathbb{R}^2| x \in  [-1,1]
\times[-1,1] \}$ and the post-condition $\postcond=\{y=f(x)\in \mathbb{R}^2~|~y_1\ge y_2\}$, and we want to prove that $\forall x. \,  x \models \precond \, \implies \, f(x) \models \postcond$.

Figure \ref{fig:demo_boundprop} shows the overall bound propagation procedure for this verification problem, where the interval $[\cdot, \cdot]$ represents the concrete value range computed for each neuron.
$z^{U}_{i}$, $z^{L}_{i}$ represent
the linear upper and lower bounding functions for nonlinear neurons, which are computed according to Equation \ref{eq:bound_fun} based on the concrete value intervals. 
Starting from the input layer, we can first compute the concrete bounds ($[-2,2]$) for $z_1$ and $z_2$ (before activation).
The bounding functions $(z^{L}_{1}$, $z^{U}_{1})$, $(z^{L}_{2}$, $z^{U}_{2})$ are then computed according to Equation \ref{eq:bound_fun}, where $\alpha=0$ is taken as the lower bounding function coefficient. 
The linear bounding functions can directly propagate to the next layer via the linear matrix transformation.
Then, by taking the minimum value of the lower bounding function and the maximum of the upper one, concrete value ranges ($[0,7]$ and $[-2,4]$) are computed for $z_3$ and $z_4$, based on which symbolic functions $(z^{L}_{3}$, $z^{U}_{3})$, $(z^{L}_{4}$, $z^{U}_{4})$ can be derived and further propagated to the output layer.
In the end, we compute the global  lower and upper bounds for $y_1$ and $y_2$, which are $[0,7]$ and $[-17.4,0]$, respectively.
From the certified bounds on the output layer, it holds that $\min(y_1) \ge \max(y_2)$
for any input $(x_1, x_2) \in [-1,1]
\times[-1,1]$. Therefore, the bound propagation method certifies that the neural network is robust in the input domain with respect to the ground-truth label $y_1$.
\end{example}

\paragraph{Abstract interpretation}
Abstract interpretation \cite{CousotC92, CousotC77} is a classic framework that can provide sound and computable finite approximations for infinite sets of behaviours. To provide sound analysis of neural networks, several works \cite{Gehr18, MirmanGV18, SinghGMPV18, Singh19} have exploited this technique to reason about safety properties. 
These methods leverage numerical abstract domains to overapproximate the inputs and compute  an over-approximation of the outputs layer by layer. 
To this end,  an abstract domain is selected to characterize the reachable output set for each layer as an abstract element.
The choice of abstract domain is essential to balance the analysis precision and scalability.
Commonly used abstract domains for neural network verification \cite{Albarghouthi21} include \textit{Interval}, \textit{Zonotope}, and \textit{Polytope}, of which the general formulations are summarized in the following (increasing in precision):  
\begin{align*}
    &\text{Interval:} \quad  \{x \in \inset | l_i \le x_i \le u_i\} \\
    &\text{Zonotope:} \, \{x \in \inset | x_i = c_{i0}+\sum_{j=1}^{m}c_{ij}\cdot \epsilon_{j}, \epsilon_j \in [-1,1]\}\\
    &\text{Polytope:} \, \{x \in \inset | x_i = c_{i0}+\sum_{j=1}^{m}c_{ij}\cdot \epsilon_{j}, F(\epsilon_1, \cdots, \epsilon_m)\}
\end{align*}
where $\epsilon_j$ ($j=1,\cdots,m$) denote $m$ generator variables. The generator variables are bounded within the interval $[-1,1]$ for zonotopes and constrained by $F$ for polytopes, where $F$ takes in the form of a convex polytope $\mathbf{c}\mathbf{x}\le \mathbf{d}$.

With the abstract domain capturing the reachable outputs of each layer, abstract transformers are defined to compute the effect of different layers on propagating the abstract element.
Affine transformers are usually supported by the underlying abstract domain, such as Zonotope and Polytope, to abstract the linear functions.
For nonlinear functions, case splitting and unifying is proposed in \cite{Gehr18} by defining the \textit{meet} and \textit{join} operators to propagate zonotope abstraction through piecewise-linear layers.
Convex approximations are adopted in \cite{Singh19} for abstract transformers of nonlinear functions where the approximation can be captured  with the proposed polyhedra abstraction.
At the end of the analysis, the abstract element of the output layer is an over-approximation of all possible concrete outputs corresponding to the input set. 
Then we can directly verify the over-approximation of the outputs against the postcondition $\postcond$, i.e., check whether the over-approximation is fully contained within $\postcond$. One drawback of this method is that the over-approximation may be quite loose.

\subsubsection{Complete methods}
Early complete verification approaches for neural networks \cite{ehlers2017formal,Katz17} encode the neural network into a set of constraints exactly and then check the satisfaction of the property with constraint solvers, e.g, SMT (Satisfiability Modulo Theory) or MILP (Mixed Integer Linear Programming) solvers. 
Since such constraint-solving methods encode the neural network in an exact way, they are able to ensure both soundness and completeness in providing certification guarantees.
One limitation is that these methods suffer from exponential complexity in the worst case. 
To address the computational intractability, Branch and Bound techniques are adopted and customized for neural network verification, where efficient incomplete methods can be exploited to speed up the bound computation.

\paragraph{SMT solver}
Reluplex \cite{Katz17} is proposed as a customized SMT solver for neural network verification.
The core idea is to extend the simplex algorithm, a standard algorithm to solve linear programming problems, with additional predicates to encode (piecewise linear) ReLU functions and transition rules (\textit{Pivot} and \textit{Update}) to handle ReLU violations.
The extended Reluplex algorithm allows variables that encode ReLU nodes to temporarily violate the ReLU constraints. 
Then, as the iteration proceeds, the solver picks variables that violate a ReLU constraint and modifies the assignment to fix the violation using \textit{Pivot} and \textit{Update} rules.
When the attempts to fix a ReLU constraint using \textit{Update} rules exceed a threshold, a ReLU splitting mechanism is applied to derive two sub-problems. Reluplex is then invoked recursively on these two sub-problems.
Compared with the eager splitting on all ReLU neurons, Reluplex proposes a splitting-on-demand strategy to reduce unnecessary splitting and limit splits to ReLU constraints that are more likely to cause violation problems.
Due to the exact encoding nature, Reluplex suffers from exponential complexity in the worst case and thus cannot scale to large neural networks.

\paragraph{MILP}
MILP-based verification methods \cite{DuttaJST18, FischettiJ18, Tjeng19} encode a neural network with piecewise-linear functions as a set of mixed integer linear constraints.
To encode the nonlinearities, they introduce an indicator decision variable $\activar$ to characterize the two statuses of unstable ReLU neurons.
An unstable ReLU neuron $\vecact=\max(\vechat,0)$ with concrete bounds $(l,u)$ can be encoded exactly using the following constraints:
\begin{align*}\label{eq:milp}
    & \vecact \ge 0, \quad \vecact \le u \cdot \activar, \\
    & \vecact \ge \vechat,  \quad
    \vecact \le \vechat - l\cdot(1-\activar), \\ 
    & \activar \in \{0,1\} 
\end{align*}
Note that the MILP constraints require the pre-computation of finite bounds for the nonlinear neurons, i.e., $(l,u)$.
It is known that the tightness of lower and upper bounds in the indicator constraints is crucial to the resolution of the MILP problem \cite{Vielma15, FischettiJ18},  and consequently, the verification efficiency. 
MIPVerify \cite{Tjeng19} thus proposes a progressive bound tightening approach to improve upon existing MILP-based verifiers.  
The algorithm starts with coarse bounds computed using efficient bound computation procedures such as \textit{Interval Arithmetic}. Bound refinement is performed only when the MILP problem can be further tightened. In such a  case, more precise but less efficient bound computation procedures, e.g., \textit{Linear Programming} (LP), are adopted to derive tighter bounds. 
This progressive bounding procedure can also be extended to other bound computation methods, such as dual optimization, to achieve a trade-off between tightness and computational complexity.

\begin{example}\label{eg:milp}
In this example, we encode the neural network, shown in Example \ref{eg:net_demo}, into the exact MILP formulation. 
The verification problem is the same as shown in Example \ref{eg:boundProp}, i.e., to determine whether $\postcond=\{y\in \mathbb{R}^2~|~y_1\ge y_2\}$ holds for all inputs in the input domain $[-1,1]\times[-1,1]$.
We encode the output property by specifying its negation, i.e., $\postcond' = \neg \postcond$. 
If there exists an instance where $\postcond'$ does hold, then a witness to $\postcond'$ is the counter-example for $\postcond$. 
If $\postcond'$ is unsatisfiable,  then the property $\postcond$ is proved.

Assume we have computed the concrete value of lower and upper bounds of $z_i$ employing efficient bound propagation techniques.
Then the neural network and the verification problem can be formulated as follows:
\begin{align}
& x_1 \ge -1, \, x_1 \le 1, \, x_2 \ge -1, x_2 \le 1 \,(\precond)\\
& \vechat_{1}=x_{1}+x_{2}, \quad \vechat_{2}=x_{1}-x_{2}, \quad \activar_1, \activar_2 \in \{0,1\}\\
    & \vecact_{1} \ge 0, \,\vecact_{1} \le 2\activar_1, \, \vecact_1 \ge \vechat_1,  \,
    \vecact_1 \le \vechat_1 + 2(1-\activar_1),\\
    & \vecact_{2} \ge 0, \,\vecact_{2} \le 2\activar_2, \, \vecact_2 \ge \vechat_2,  \,
    \vecact_2 \le \vechat_2 + 2(1-\activar_2),\\
     & \vechat_{3} = \vecact_{1} + 3\vecact_{2}, \quad
     \vechat_{4} = -\vecact_{1} +2 \vecact_{2}, \quad \activar_4 \in \{0,1\}\\
    & \vecact_{3} = \vechat_3, \,\text{(stable neuron)}  \\
    & \vecact_{4} \ge 0, \,\vecact_{4} \le 4\activar_4, \, \vecact_4 \ge \vechat_4,  \,
    \vecact_4 \le \vechat_4 + 2(1-\activar_4),\\
     & y_1 = \vecact_{3}, \quad
     \quad y_2 = -2\vecact_{3} - \vecact_{4}\\
     & y_1 < y_2 \, (\neg \postcond)
\end{align}
The lower and upper bounds $l_i$ and $u_i$ ($i\in\{1,2,3,4\}$) are derived as shown in Example \ref{eg:boundProp}.
The binary variables $\activar_i$ ($i\in\{1,2,4\}$) are introduced to indicate the status of unstable ReLUs and it holds that
$\activar_i = 0 \Leftrightarrow \vecact_i = 0$ and $\activar_i = 1 \Leftrightarrow \vecact_i = \vechat_i$. 
Checking the feasibility of the above model using MILP solvers (e.g., Gurobi) will return infeasible, thus proving the original property.
\end{example}
\paragraph{Branch and Bound}
To improve the scalability of verification algorithms to larger neural networks, a branch and bound framework (BaB) \cite{BunelLTTKK20} has been proposed. 
The BaB framework mainly consists of two components: a branching method that splits the original verification problem into multiple subproblems and a bounding method to compute the upper and lower bounds of the subproblems. 
This modularized design provides a unifying formulation paradigm for different verifiers, with the main difference lying in the splitting function and the bounding method. For example, the verifier
\textit{ReluVal} \cite{WangPWYJ18} performs splitting on the input domain according to sensitivity analysis, e.g., input-output gradient information, and computes bounds using symbolic interval propagation. 
The aforementioned SMT-based verifier \textit{Reluplex} \cite{Katz17} performs splitting on ReLU neurons guided by the violation frequency of the ReLU constraints and computes the bounds on the relaxed problems by dropping some constraints on the nonlinearities (which yields an over-approximation of the constraint optimization problems).

To further improve neural network verification,  the BaB method introduces two new branching strategies: BaBSB for branching on input domains and BaBSR for branching on ReLU neurons.
Both branching methods adopt a similar heuristic to decide which dimension or ReLU neuron to split on.  BaBSB computes a rough estimate of the improvement on the bounds obtained with regard to every input dimension, where the estimation makes the split decision be set more efficiently. On the other hand, BaBSR estimates the bound improvement with regard to each unfixed ReLU neuron by computing the ReLU scores. The bounding methods resort to LP solvers to tighten the intermediate bounds on the subdomains or use more computationally efficient methods such as Interval Arithmetic.

\subsection{Backward Analysis Methods}
Backward analysis methods for neural networks, also known as preimage generation or inverse abstraction, aim at computing the input set that will lead the neural network to a target set, e.g., a safe or unsafe region. 
They complement the forward analysis methods, which may result in over-approximated bounds worsening as the computation progresses through the layers of the network.
In the following, we categorize the representative approaches broadly into two groups: \textit{exact} and \textit{approximate} methods.

\subsubsection{Exact methods}
Exact backward analysis methods reason about the preimage of a target output set by encoding the neural network behaviours in an exact manner. These methods are able to compute the exact symbolic representation of the preimage for different output properties. One limitation suffered by these methods is that they can only process neural networks with piecewise-linear activation functions (e.g., $\mathit{ReLU}$), as they aim at an exhaustive decomposition of the non-convex function (the neural network) into a set of linear functions. 
The preimage (input set) for a target output set with regard to a neural network $f$ is characterized as a union of polytopes, where the mapping functions are completely linear on each subregion.
\paragraph{Exact preimage}
The exact preimage generation method \cite{Matoba20Exact} complements the forward analysis methods to reason about the inputs that lead to target outputs. The algorithm computes the exact preimage by relying on two elementary properties: (1) preimage of the composite functions is the reversed composition of preimages for each layer, i.e., $(f^{(L)} \circ \cdots \circ f^{(1)})^{-1}=(f^{(1)})^{-1}\circ \cdots \circ (f^{(L)})^{-1}$, and (2) preimage of a union set can be built up from the preimages of each subset in the union, i.e., $f^{-1}(\cup_{j}S_j)=\cup_j f^{-1}(s_j)$. This method assumes that the output set, e.g., a safe region, can be formulated as a polytope (intersection of half-planes) $\{y \in \outset | Ay-b \le 0\}$. It then propagates the polytope backwards through the layers.

For linear layers, the preimage is computed by applying the linear operations corresponding to the layer.
Suppose we have a linear mapping in the form of $y=Wz+a$, then the preimage of the output polytope under this linear operation can be formulated as
$\{z \in \mathbb{R}^{n_{L-1}} | AWz+(Aa-b)\le 0\}$.
For nonlinear layers, the algorithm restricts the backward propagation to a subset where the activation pattern of the ReLU neurons is fixed.
Let $s(z)$ denote the activation status vector of the nonlinear neurons where $s(z)_j=1$ if $z_j \ge 0$ and $s(z)_j=0$ otherwise.
A diagonal matrix $diag(s(z))$ is introduced to restrict to a fixed activation pattern, on which only linear computation is required to compute the preimage subset. The exact preimage can then be computed by taking the union of each partition (preimage property (2)). 
\begin{align*}
&\quad \mathit{ReLU}^{-1}(\{y \in \outset|Ay-b \le 0\})\\
&= \bigcup_{s\in \{0,1\}^{n_{i}}} \{z \in \mathbb{R}^{n_i} | Adiag(s)z -b \le 0, -diag(s)z\le 0, \\
& \, \quad \quad \quad \quad \quad diag(1-s)z \le 0\}
\end{align*}
\begin{figure}[t]
    \begin{subfigure}[t]{0.48\columnwidth}
    \includegraphics[width=\linewidth]{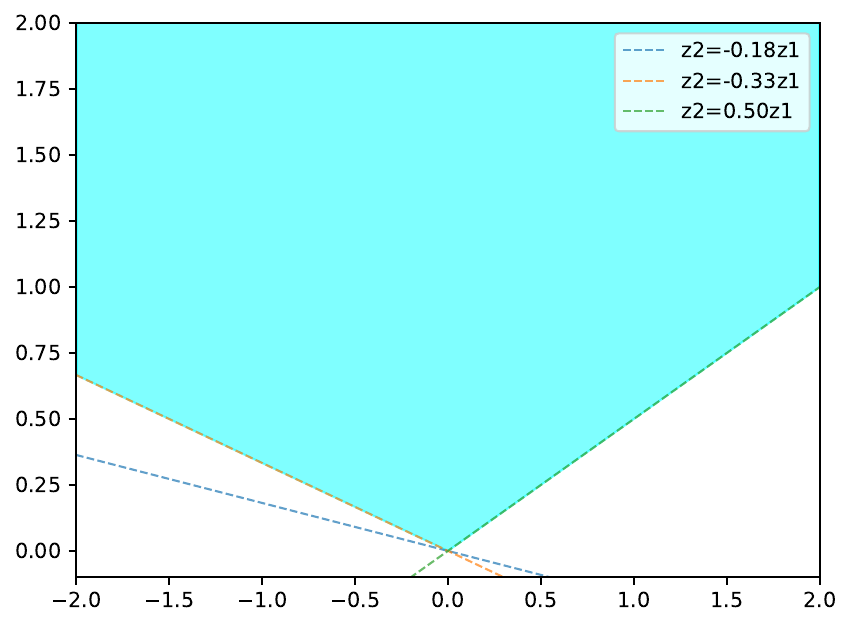}
    \end{subfigure}
    \hspace{\fill}
    \begin{subfigure}[t]{0.48\columnwidth}
    \includegraphics[width=\linewidth]{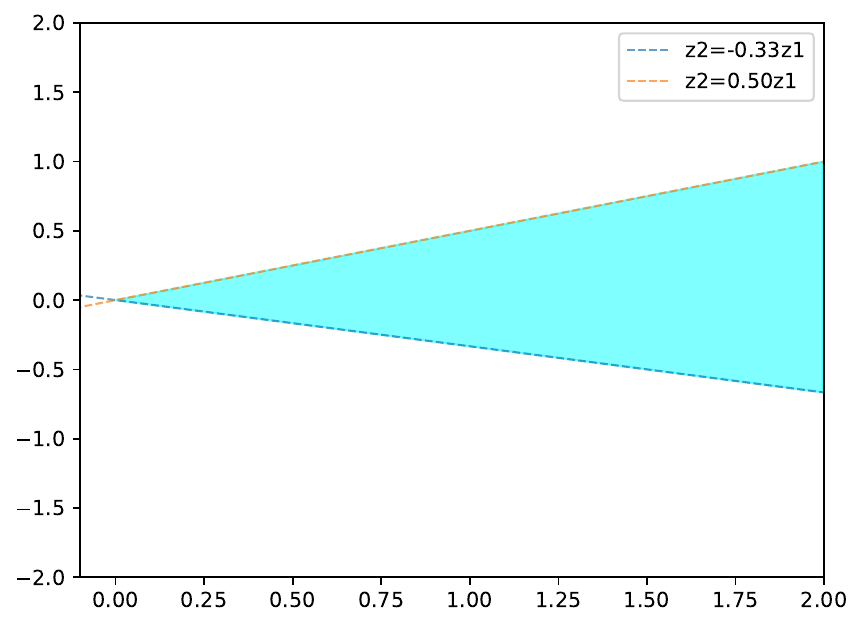}
    \end{subfigure}
    \bigskip
    \begin{subfigure}[t]{0.48\columnwidth}
    \includegraphics[width=\linewidth]{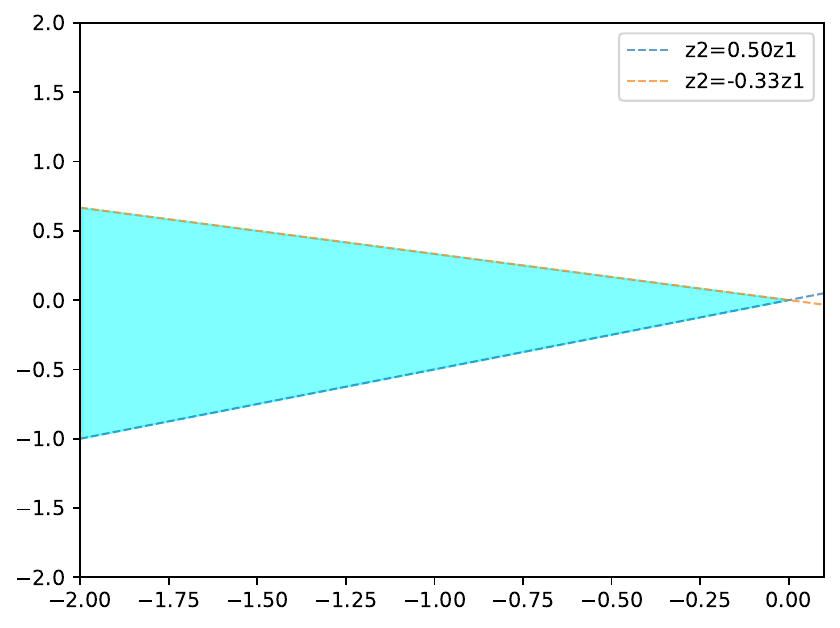}
    \end{subfigure}
    \hspace{\fill}
    \begin{subfigure}[t]{0.48\columnwidth}
    \includegraphics[width=\linewidth]{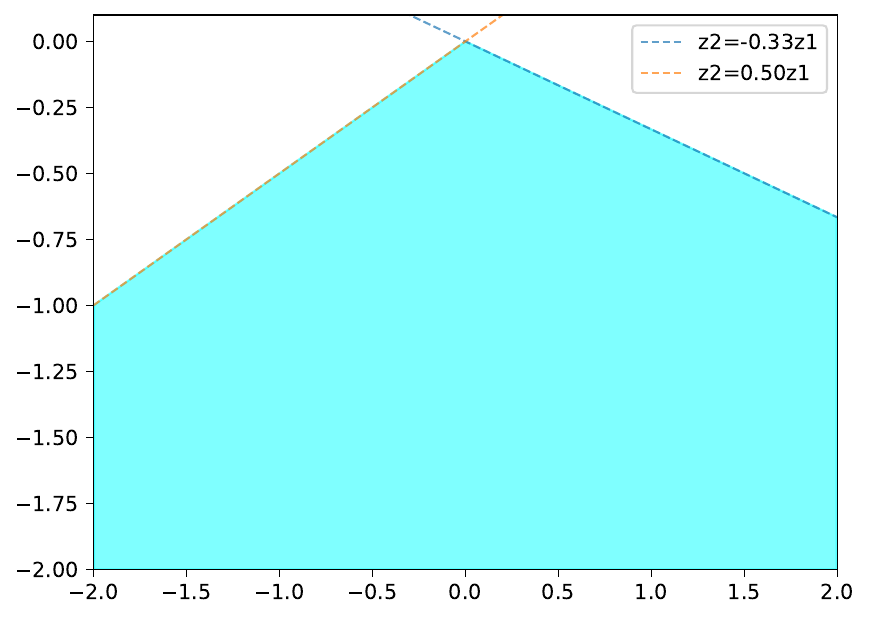}
    \end{subfigure}
    \caption{Preimage polytopes by exact method.}
    \label{fig:demo_exact_preimage}
\end{figure}
\begin{example}\label{eg:exact_preimage}
    In this example, we     consider the same verification problem as in Example \ref{eg:boundProp} and \ref{eg:milp}, but from the backward perspective. 
    Preimage analysis aims to investigate whether the input region $[-1,1]\times[-1,1]$, which is expected to result in decision $y_1$, fails the safety check.
    We first formulate the target output region as a polytope.
    Since we only have two labels, the output constraint is, therefore, a single half-plane encoded as
    $\{y \in \mathbb{R}^2| y_1 - y_2 \ge 0\}$.
    We then proceed to compute the preimage  of the target polytope under the linear mapping (from the $2^{\mathit{nd}}$ hidden layer to the output layer), of which the result is
    $\{(z_3, z_4) \in \mathbb{R}^2~|~ 3z_3 + z_4 \ge 0\}$.
    Next, preimage computation for ReLU starts with partitioning the neuron vector  space $\mathbb{R}^2$ into $2^{2}$ sets where, for each subset, the status of nonlinear neurons is fixed and preimage computation proceeds similarly to the linear mapping.
    The partition leads to four result polytopes.

Figure \ref{fig:demo_exact_preimage} shows the result of four preimage polytopes derived in the two-dimensional space $(z_1, z_2)$. 
As an example, the preimage polytope derived corresponding to the partition where both neurons are active is (upper left of Figure \ref{fig:demo_exact_preimage}):
\begin{align*}
    &\{\vecact^{(1)} \in \mathbb{R}^2\ : A^{(1)}\vecact^{(1)} \ge 0\} \quad
    \text{where}\\
    &  A^{(1)} = \left[
 \begin{matrix}
    2 & 11\\
   1 & 3\\
   -1 & 2\\
  \end{matrix}
  \right], \quad \vecact^{(1)}= \left(
 \begin{matrix}
    \vecact_1 \\
   \vecact_2
  \end{matrix}
  \right)
\end{align*}
The other three polytopes are derived in the same way.
The four preimage polytopes are then partitioned further into 16 polytopes to characterize the exact preimage of the input layer.
Note that the combination of the four polytopes actually covers the hidden vector space $[-2,2]\times[-2,2]$, and the resulting preimage polytopes on the input layer cover the region $[-1,1]\times[-1,1]$, which certifies that the correct decision is taken for the entire region under investigation.

\end{example}
\paragraph{SyReNN}
SyReNN is proposed in \cite{SotoudehTT23} to compute the symbolic representation of a neural network so as to understand and analyze its behaviours. It targets \textit{low-dimensional} input subspaces and computes their exact symbolic partitioning, on which the mapping function is completely linear. This methodological design is also referred to as neural network decomposition. We classify it as a backward analysis method, as this method provides a symbolic representation in the input space.
SyReNN focuses on neural networks with piecewise-linear activation functions.
This restriction enables a precise characterisation of the input space $X$ as a finite set of polytopes $\{X_1, \cdots, X_n\}$.  Within each input polytope $X_i$, the neural network is equivalent to a linear function. By means of such a symbolic representation, safety verification is reduced to checking whether the vertices of every bounded convex polytope $X_i$ satisfy the output property.

To compute the symbolic decomposition on the input domain, this algorithm starts with the trivial partition $X$ and derives the linear partitions layer by layer. 
Given the partition hyperplanes of the nonlinear layer $i$, e.g., $z_1=0, z_2=0, \cdots, z_{n_i}=0$ with $n_i$ ReLUs, and the symbolic representation $\hat{f}_{i-1}$ (a set of polytopes) computed until layer  $i-1$,
$\hat{f}_{i}$ is computed by recursively partitioning the current polytopes based on the newly-added hyperplanes.
For example, given a polytope $Z_{i-1}$, if an orthant boundary (e.g., hyperplane $z_i=0$) is hit when traversing the boundary of $Z_{i-1}$, then $Z_{i-1}$ is further partition into $Z_{i-1,1}$ and $Z_{i-1,2}$ which lie on the opposite sides of the hyperplane.
This procedure terminates until all resulting polytopes lie within a completely linear region of the neural network $f$. 

\subsubsection{Approximate methods}
Exact methods for preimage analysis suffer from exponential complexity in the worst case. Similarly to the development of incomplete verifiers, preimage approximation techniques begin to emerge by leveraging different approximation (relaxation) techniques. 
They compute a symbolic approximation of the preimages to bypass the intractability of computing exact preimage representations. Computational efficiency and scalability can be greatly improved with the sacrifice of precision.

\paragraph{Symbolic interpolation}
Symbolic interpolation \cite{Albarghouthi13interpolants} has been used for program verification and SMT solving. 
To compute provable preimage approximations, \cite{Dathathri19Inverse} leverages interpolants, especially those with simple structures, and computes preimages from the output space through hidden layers to the input space.  
The generated approximations can then be applied to reason about the properties of the neural network itself. 
For example, in the case that a desired property (a target output set) $Y$ should be satisfied when starting from a certain input set $X$, an under-approximation $\underline{X}$ of the preimage for $Y$ can be computed. Then the property can be verified by checking whether $\underline{X} \rightarrow X$ holds. 

\cite{Albarghouthi13interpolants} proposes an algorithm to compute the preimage approximation by iterating backwards through the layers. It encodes the neural network as constraints in the theory of \textit{quantifier-free linear rational arithmetic} (QFLRA) and requires the output set to be encoded as a Boolean combination of atoms in the form of half-spaces. 
Suppose we now focus on deriving preimage 
over-approximations of the target output set $Y$.
The algorithm starts by computing the (overapproximated) set of inputs to the last layer, denoted as $p_{L}^{f, Y}$, which leads to the output set $Y$, i.e., $f^{(L)}(p_{L}^{f, Y}) \models Y$.
The algorithm then iteratively computes preimages of the other layers that satisfy $p_{i}^{f,Y}=\{z ~|~ f^{(i)}(z) \models p_{i+1}^{f,Y}\}$.
This procedure leverages sampling techniques to construct a set of points mapped to the complement of $Y$, which are used to tighten the over-approximations.
The algorithm relies on Craig's Interpolation theorem to guarantee the existence of an (over- and under-) approximation. It also leverages the bound propagation framework to compute a bounded domain on each layer, which speeds up the interpolation condition checking. 

\paragraph{Inverse bounding}

\cite{Kotha23BoundPreimage} points out two important applications based on preimage analysis: safety verification for dynamical systems and out-of-distribution input detection. Motivated by these use cases, an inverse bound propagation method is proposed to compute the over-approximation of the preimage. Bound propagation has been widely employed to build efficient verifiers in the forward direction (certified output bound computation). Compared with forward analysis, it is challenging to adopt bound propagation methods directly to compute tight intermediate bounds, and thus difficult to compute tight over-approximations. This is because,  for the inverse problem, the constraints on the input are quite loose and even unbounded in some control applications.
Simply applying the bound propagation procedure will not lead to useful intermediate bounds, which further impacts the tightness of the symbolic relaxation on nonlinear neurons. 

Given this, an inverse propagation algorithm is proposed in \cite{Kotha23BoundPreimage} to compute a convex over-approximation of the preimage represented by a set of cutting planes.
It first transforms the preimage over-approximation problem to a constrained optimization problem over the preimage and further relaxes it to Lagrangian dual optimization.
To tighten the preimage and intermediate bounds, they introduce a dual variable with respect to the output constraints and tighten these bounds iteratively, leveraging standard gradient ascent algorithm.

\paragraph{Preimage approximation}
Motivated by the practical needs of global robustness analysis \cite{ruan2019hamming, Leino21globalrobust,wang2021statistically} and quantitative verification \cite{baluta2021quantitative,yang2021quantitative}, an anytime algorithm is proposed in \cite{ZWK23} to compute provable preimage approximation.
The generated preimage is further applied to verify 
quantitative properties of neural networks, which is defined by the relative proportion of the approximated preimage volume against the input domain under analysis, formally defined as follows.
\begin{definition}
    Given a neural network $\nn: \inset \rightarrow \outset$, a measurable input set with non-zero measure (volume) $X \subseteq \inset$, a measurable output set $Y \subseteq \outset$, and a rational proportion $p \in [0, 1]$, the neural network satisfies the quantitative property $(X, Y, p)$ if $\frac{\volume(\nn^{-1}_{X}(Y))}{\volume(X)} \geq p$.
\end{definition}

This approach targets safety properties that can be represented as polytopes and characterizes preimage under-approximation using a disjoint union of polytopes.
To avoid the intractability of the exact preimage generation method, convex relaxation is used to derive sound under-approximations.
However, one challenge is that the generated preimage under-approximation  can be quite conservative when reasoning about properties in large input spaces with relaxation errors accumulated through each layer.
To refine the preimage abstraction, a global branching method is introduced to derive tighter approximations on the input subregions.
This procedure proposes a (sub-)domain search strategy prioritizing partitioning on most uncovered subregions and a greedy splitting rule leveraging GPU parallelization to achieve better per-iteration improvement.
To further reduce the relaxation errors, this method formulates the approximation problem as an optimization problem on the preimage polytope volume. Then it proposes a differentiable relaxation to optimize bounding parameters using projected gradient descent. 

\begin{figure}[t]
    \begin{subfigure}[t]{0.48\columnwidth}
    \includegraphics[width=\linewidth]{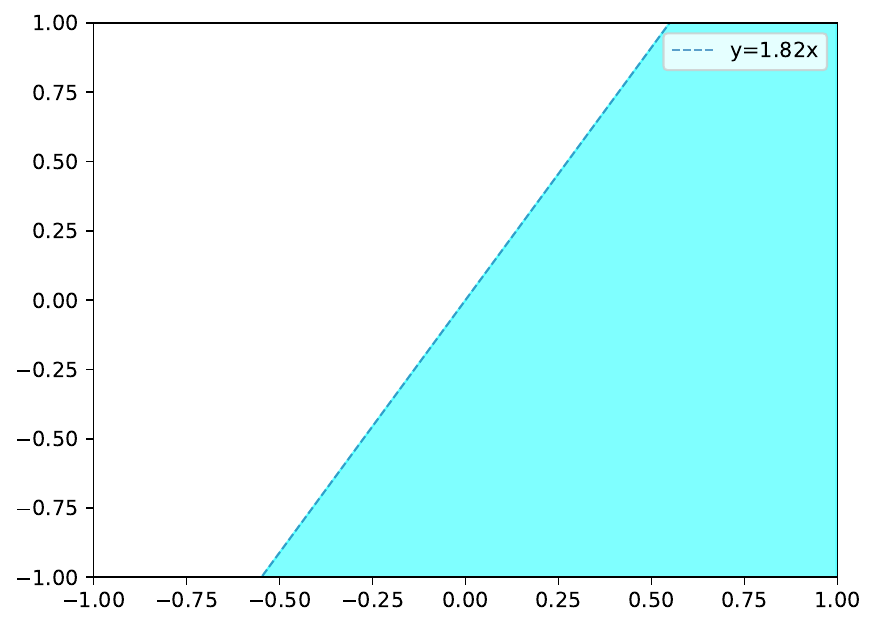}
    \end{subfigure}
    \hspace{\fill}
    \begin{subfigure}[t]{0.48\columnwidth}
    \includegraphics[width=\linewidth]{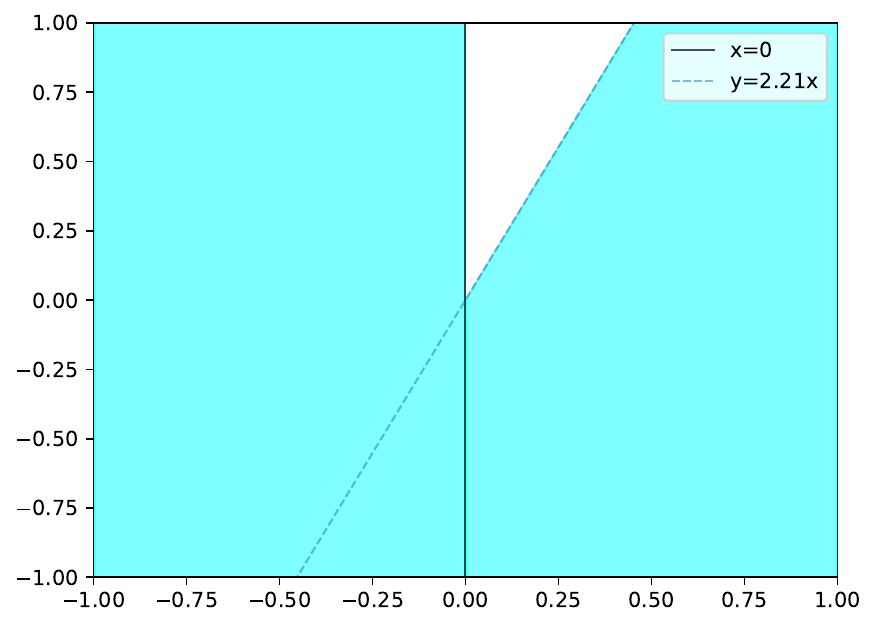}
    \end{subfigure}
    \caption{Preimage approximation.}
    \label{fig:demo_approx_preimage}
\end{figure}
\begin{example}\label{eg:preimage_approx}
In this example, we demonstrate how to construct a provable preimage (under-)approximation for the target output region, and apply it to quantitative analysis of the verification problem shown in previous examples. 
Consider the quantitative property with input set $\precond= \{x \in \mathbb{R}^{2}~|~x \in [-1,1]^2\}$, output set $\postcond=\{y \in \mathbb{R}^{2}~|~  y_1-y_2 \ge 0\}$, and quantitative proportion $p=0.9$.
We apply the preimage approximation algorithm to verify this property.
Figure \ref{fig:demo_approx_preimage} presents the computed preimage before (left) and after one-iteration  refinement (right). Note that the partition is performed w.r.t. input $x_1$, which results in two polytopes for the subregions.
We compute the exact volume ratio of the refined under-approximation against the input set.
The quantitative proportion reached with the refinement is 94.3\%, which verifies the quantitative property.
\end{example}

\section{Application Examples}\label{sect-IV}
In this section we provide a selection of experimental results and lessons learnt from applying formal verification and certification approaches described in the previous section to neural network models drawn from a range of classification problems. These include image and video recognition, automated decisions in finance and text classification. In addition to adversarial robustness of the models, we demonstrate certification of individual fairness of automated decisions and discuss robust explanations.

\subsection{MSR-based Certification for Images and Videos}
\begin{figure}[t]
	\centering
	\includegraphics[width=1\linewidth]{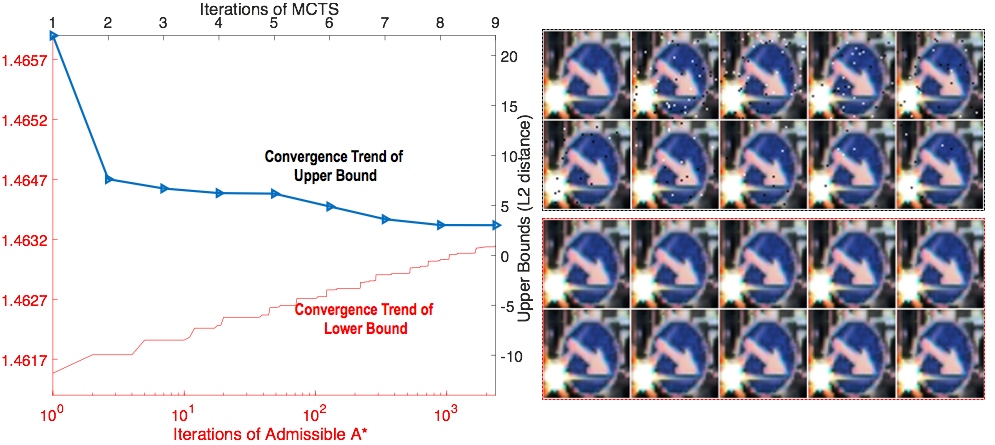}
	\caption{Convergence of maximum safe radius computed using the game-based method for a traffic sign image from the GTSRB dataset originally classified as ``keep right''. Left: The convergence trends of the upper bound obtained with Monte Carlo Tree Search and the lower bound with Admissible A*. Right: unsafe images (top two rows) and certified safe images (bottom two rows). 
 Figure taken from \cite{Wu2020}.} 
	\label{fig:converge_coop_gtsrb}
\end{figure}

The game-based method  \cite{Wu2020} has been applied to analyse and certify the robustness of image classification models to adversarial perturbations with respect to the maximal safe radius, working with a range of feature extraction methods and distance metrics. Figure \ref{fig:converge_coop_gtsrb} shows a typical outcome of such analysis, with converging lower and upper MSR bounds for an image of a traffic sign for $l_2$ distance and features extracted from the latent representation computed by a convolutional neural network (CNN) model. It can be seen that the image is certified safe for adversarial perturbations of up to 1.463 in $l_2$ distance, which is some distance away from the best upper bound at approx. 3, but can be improved with more iterations since the method is anytime. 

\begin{figure}[t]
    \centering
    \includegraphics[width=0.24\linewidth]{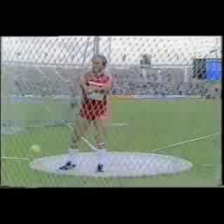}
    \includegraphics[width=0.24\linewidth]{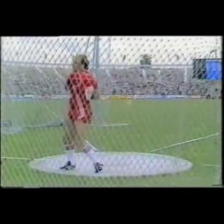}
    \includegraphics[width=0.24\linewidth]{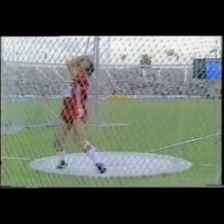}
    \includegraphics[width=0.24\linewidth]{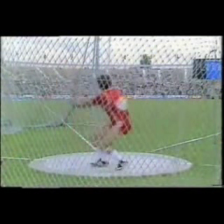}
    \\
    \includegraphics[width=0.24\linewidth]{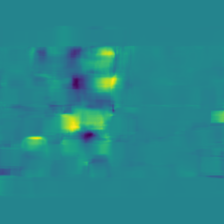}
    \includegraphics[width=0.24\linewidth]{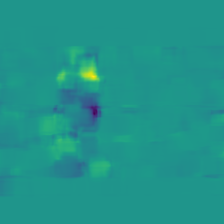}
    \includegraphics[width=0.24\linewidth]{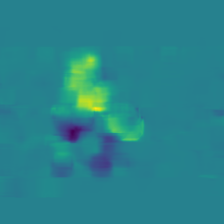}
    \includegraphics[width=0.24\linewidth]{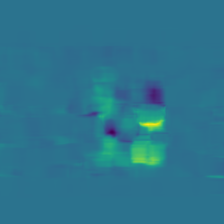}
    \\
    \includegraphics[width=0.24\linewidth]{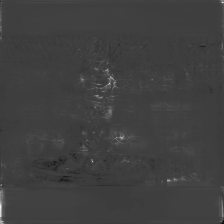}
    \includegraphics[width=0.24\linewidth]{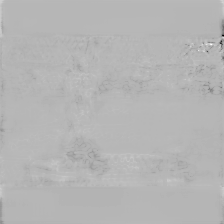}
    \includegraphics[width=0.24\linewidth]{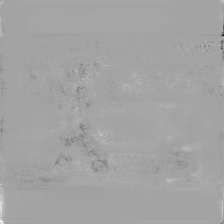}
    \includegraphics[width=0.24\linewidth]{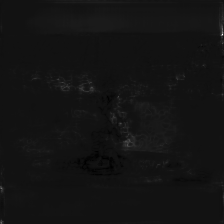}
    \\
    \includegraphics[width=0.24\linewidth]{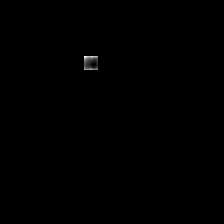}
    \includegraphics[width=0.24\linewidth]{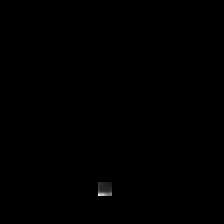}
    \includegraphics[width=0.24\linewidth]{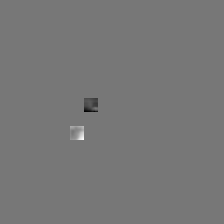}
    \includegraphics[width=0.24\linewidth]{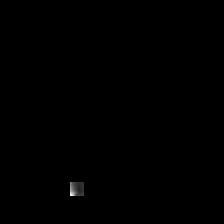}
    \caption{Shown in top row are sampled frames
    of a $\mathsf{Hammer Throw}$ video and the corresponding optical flows are in the 2nd row. Unsafe perturbations of flows are in 3rd row and safe in 4th. Figure taken from \cite{Wu2019}.
    }
    \label{fig:perturbations_HammerThrow}
\end{figure}

An extension of the game-based method was developed in~\cite{Wu2019} to provide MSR-based certification for videos, and specifically for neural network models consisting of a CNN to perform feature extraction and a recurrent neural network (RNN) to process video frames.  Adversarial perturbations were defined with respect to optical flow, and the algorithmic techniques involve tensor-based computation. Examples of safe and unsafe perturbations are shown in Figure~\ref{fig:perturbations_HammerThrow}, and convergence trends for lower and upper bounds similar to those in Figure \ref{fig:converge_coop_gtsrb} can be observed.

\subsection{Robustness of Language Models}
As an example of application of convex relaxation tools (variants of CROWN \cite{zhang2018crown}), we mention the study of \cite{la2020assessing}, which aims to assess the robustness of Natural Language Processing tasks (sentiment analysis and text classification) to word substitution. It was reported that standard fully connected (FC) and CNN models are very brittle to such perturbations, which may make their certification unworkable. \cite{LK22} critiqued the appropriateness of the classical concept of adversarial robustness defined in terms of word substitution in the context of NLP models.
It was observed in an empirical study that models trained to be robust in the classical sense, for example, trained using interval bound propagation (IBP), lack robustness to syntax/semantic manipulations. It was then argued in \cite{LK22} that a {\em semantic} notion of robustness that better captures linguistic phenomena such as shallow negation and sarcasm is needed for language models, where a framework based on templates was developed for evaluation of semantic robustness.

\subsection{Robust Explanations for Language Models}
\begin{figure*}
\centering
\includegraphics[width=\linewidth]{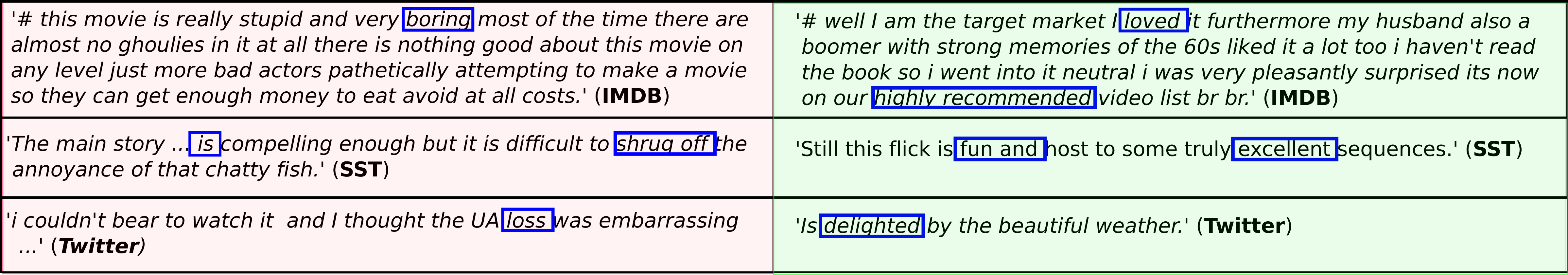}\caption{
Optimal robust explanations (highlighted in blue) for IMDB, SST and Twitter datasets (all the texts are correctly classified).
Figure taken from \cite{LZM+21}. 
}
\label{fig:good-OREs}
\end{figure*}

\begin{figure}
\centering
\includegraphics[height=3.8cm,width=\linewidth]{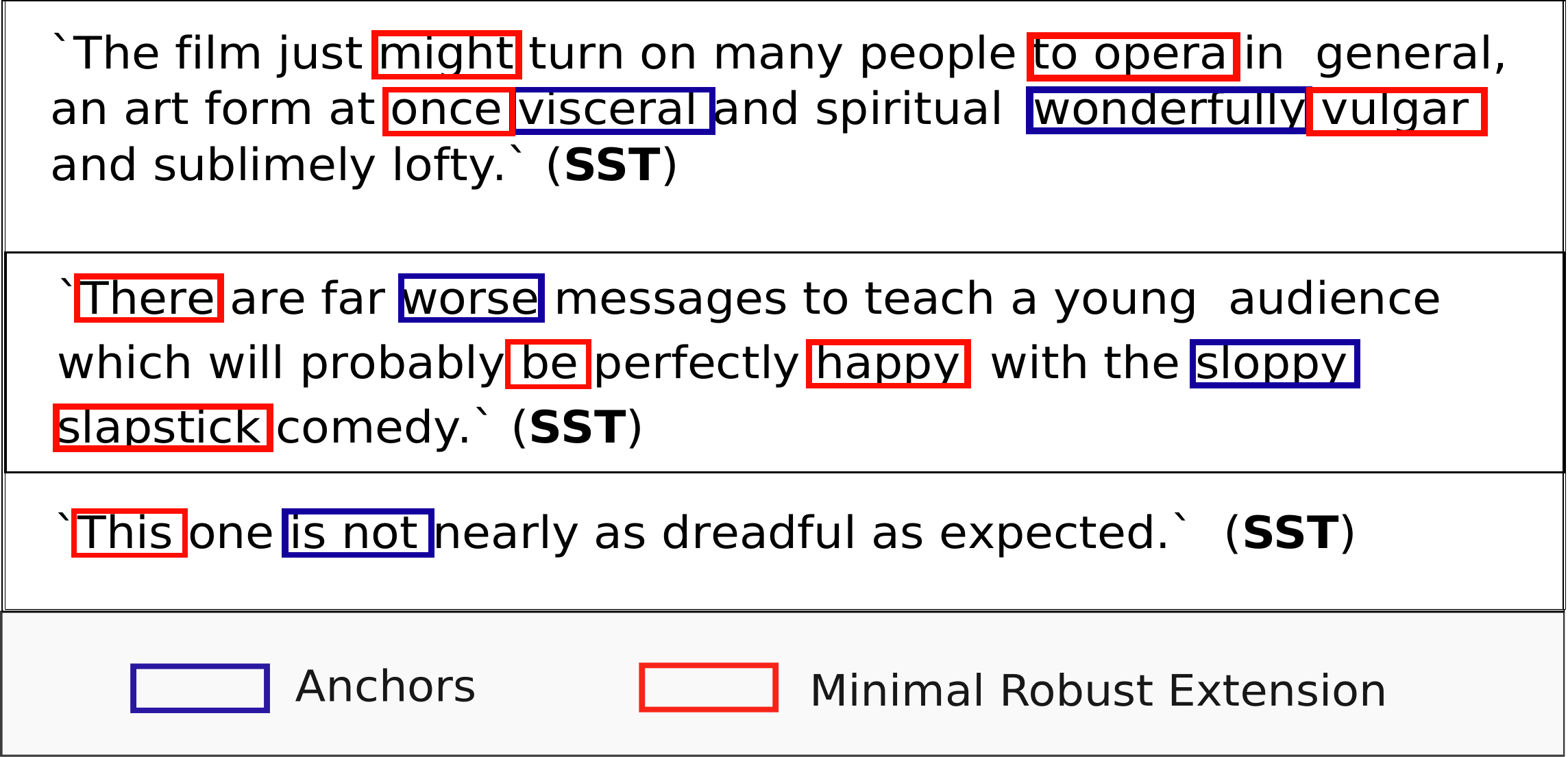}\caption{
Examples of Anchors explanations (in blue) along with the minimal extension required to make them robust (in red).
Figure taken from \cite{LZM+21}.
}
\label{fig:anchors-completion}
\end{figure}

Explainability of language models was studied in~\cite{LZM+21}, with a focus on robust optimal explanations that imply the model prediction. 
Figure~\ref{fig:good-OREs} shows examples of high-quality robust optimal explanations (using the minimum length of explanation as the cost function). In contrast, heuristic explanations such as integrated gradients or Anchors my lack of robustness, but it is possible to repair non-robust Anchors explanations by minimally extending them, see Figure~\ref{fig:anchors-completion}. 

\subsection{Fairness Certification Using MILP}
\begin{figure*}[ht]
\centering
  \centering
  \includegraphics[width=0.8\textwidth]{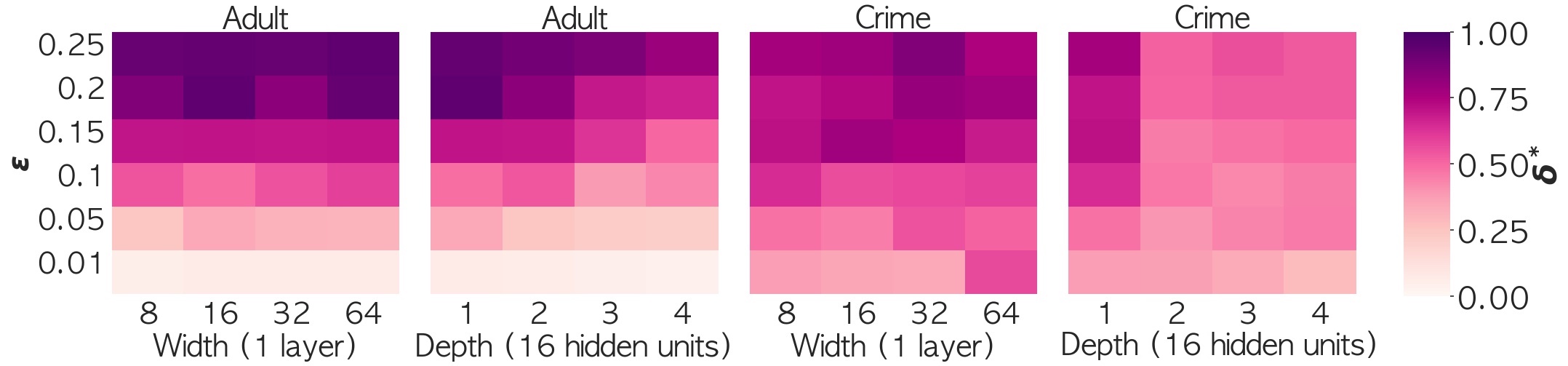}\label{fig:verification-mahalanobis} \\
  \includegraphics[width=0.8\textwidth]{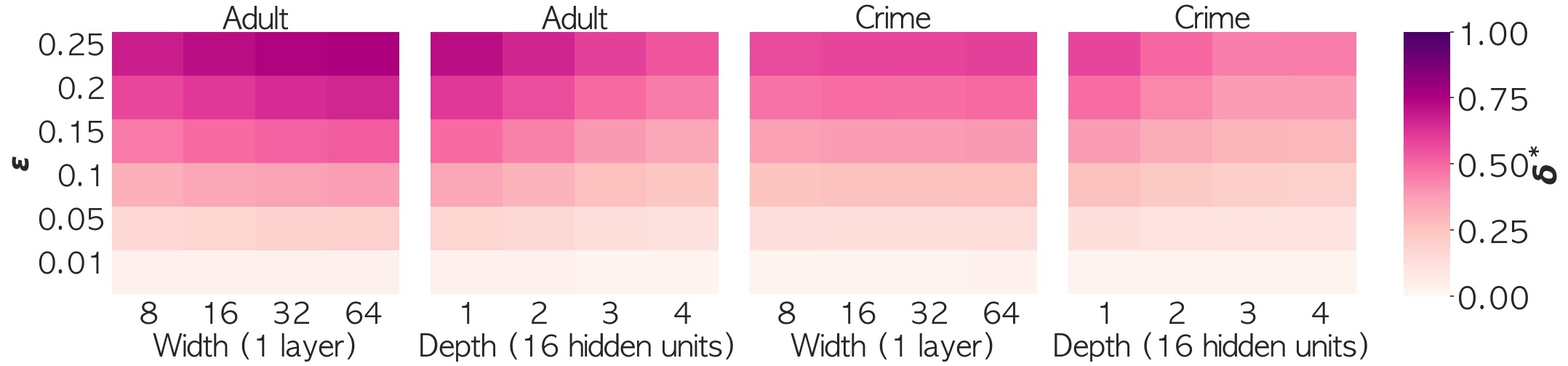}\label{fig:verification-weighted-lp}
  \caption{Certified bounds on individual fairness ($\delta_*$) for different architecture parameters (widths and depths) and maximum similarity ($\epsilon$) for the Adult and the Crime datasets. Similarity metrics used are Mahalanobis (top row) and weighted $\ell_{\infty}$ metric (bottom row).
  Figure taken from \cite{BPW+22}.
  }
  \label{fig:verification-fig}
\end{figure*}

\cite{BPW+22} developed methods for certification of individual fairness of automated decisions, defined, given a neural network and a similarity metric learnt from data, as requiring that the output difference between any pair of $\epsilon$-similar individuals is
bounded by a maximum decision tolerance
$\delta \geq 0$. Working with a range of similarity metrics, including Mahalanobis distance, a MILP-based method was developed not only to compute certified bounds on individual fairness, but also to train certifiably fair models. 
The computed certified bounds $\delta_*$ are plotted in Figure \ref{fig:verification-fig} for the Adult and the Crime benchmarks. 
Each heat map depicts the variation of $\delta_*$ as a function of $\epsilon$ and the NN architecture.
It can be observed that increasing $\epsilon$ correlates with an increase in the values for $\delta_*$, as higher values of $\epsilon$ allow for greater feature changes.

\section{Future Challenges}\label{sect-V}
Formal verification and certification of neural network models has made steady progress in recent years, with several tools released to the community and an established tool competition~\cite{VNNComp}. 
Nevertheless, considerable scientific and methodological progress is needed before these tools are adopted by developers. Below we outline a number of research challenges.

\paragraph{Beyond $\ell_p$-norm robustness} 
The vast majority of robustness evaluation frameworks consider bounded $\ell_p$-norm perturbations. While these suffice as proxies for minor visual image perturbations, real-world tasks rely on similarity measures, for example cosine similarity for word embeddings or Mahalanobis distance for images. It is desirable to define measures and certification algorithms for semantic robustness, which considers such similarity measures as first class citizens, and works with perturbations that reflect visual or geometric aspects characteristic of the application, such as object movement or lighting conditions. More generally, robustness evaluation frameworks for more complex properties induced by the use cases will be needed.

\paragraph{Beyond supervised robustness}
Existing robustness formulations focus on the supervised learning setting. However, collecting and labelling large datasets that are necessary to ensure the high robustness performance needed in safety-critical applications is costly and may not be feasible for use cases such as autonomous driving. Instead, it is desirable to formulate robustness measures and evaluation frameworks directly in some appropriate semi-supervised, or even unsupervised, setting, where the definition of robustness needs to focus on the quality of the learned representations rather than classification (prediction) because of the lack of labels. This may involve working with similarity measures such as Mahalanobis distance and will be challenging both theoretically and computationally to achieve provable robustness guarantees. 

\paragraph{Scalability in network width and depth}
Despite much progress, the scalability of robustness certification and evaluation frameworks remains limited to low-dimensional models. In order to apply certification to realistic use cases (such as object detection) will necessitate significant improvements with respect to input dimensionality and network depth, as well as the types of activation functions that can be handled.

\paragraph{Efficiency and precision trade-off}
Robustness certifications and evaluation involves a variety of methods, including exact, approximate and statistical. While exact methods offer completeness, trading off exact precision for approximate bounding results in more efficient any time methods, and completeness can be recovered by combining fast approximate methods such as convex relaxation with branch-and-bound computation. Statistical methods provide estimates of robustness that may be unsound but fast and, in many cases, sufficient for the application being considered. 

\paragraph{Compositionality and modularity of AI systems}
Certification tools that have been developed to date are monolithic, which matches the monolithic structure of the vast majority of neural network models. Yet, similarly to safety-critical systems, it is anticipated that better structuring of models and tools is likely to improve their reliability and maintainability. Therefore, modularity, compositionality and, in particular, assume-guarantee compositional frameworks, are desirable future directions.

\paragraph{Calibrating uncertainty}
It is recognised that deterministic neural networks can be overconfident in their decisions, and instead a variant known as Bayesian neural networks (BNNs), which admits a distribution over the weights and provides outputs in the form of the posterior distribution, is preferred, as it allows for a principled means to return an uncertainty measure alongside the network output. BNN certification methodologies are much more complex than for deterministic NNs, and still in early stages of development, including uncertainty quantification \cite{michelmore2019uncertainty}, computing lower bounds on safety probability \cite{wicker20} and certifiable adversarial robustness \cite{wicker2021certification}. Unfortunately, the methods do not scale beyond small networks and standard Bayesian inference tends to underestimate uncertainty.

\paragraph{Robust learning}
A drawback of certification as presented in this paper is that it pertains to trained models, and if the model fails certification, it is not clear how it can be repaired, and expensive retraining may be needed. A natural question then arises as to whether one can learn a model that is guaranteed to be robust. Building on the positive and negative theoretical results in the case of robust learning against evasion attacks~\cite{gourdeau2019hardness,gourdeau2021hardness,gourdeau2022sample,gourdeau2022when}, it would be interesting to generalise these results to neural network models and development of implementable frameworks that can provide provable guarantees on robustness.

\section{Conclusion}\label{sect-VI}
We have provided a brief overview of formal verification approaches that can be employed to certify neural network models at test time, to train certifiably robust or fair models, and to provide meaningful explanations for network predictions. The methods can be categorised into forward and backward analysis, and involve techniques such as search, bound propagation, constraint solving and abstract interpretation.
Both forward and backward analysis have the potential to support more complex verification properties, which have been little explored to date. 
Empirical results obtained on a range of standard benchmarks show that neural network models are often brittle to adversarial perturbations, but verification approaches can be used to strengthen their robustness and compute certification guarantees, thus improving trustworthiness of AI decisions. 

\section*{Acknowledgments}
This project received funding from the ERC under the European Union’s Horizon 2020 research and innovation programme (FUN2MODEL, grant agreement No.~834115) and ELSA: European Lighthouse on Secure and Safe AI project (grant agreement
No. 101070617 under UK guarantee).

\bibliographystyle{IEEEtran}
\bibliography{ref,biblio,biblio2}

\begin{thebibliography}{10}
\providecommand{\url}[1]{#1}
\csname url@samestyle\endcsname
\providecommand{\newblock}{\relax}
\providecommand{\bibinfo}[2]{#2}
\providecommand{\BIBentrySTDinterwordspacing}{\spaceskip=0pt\relax}
\providecommand{\BIBentryALTinterwordstretchfactor}{4}
\providecommand{\BIBentryALTinterwordspacing}{\spaceskip=\fontdimen2\font plus
\BIBentryALTinterwordstretchfactor\fontdimen3\font minus \fontdimen4\font\relax}
\providecommand{\BIBforeignlanguage}[2]{{%
\expandafter\ifx\csname l@#1\endcsname\relax
\typeout{** WARNING: IEEEtran.bst: No hyphenation pattern has been}%
\typeout{** loaded for the language `#1'. Using the pattern for}%
\typeout{** the default language instead.}%
\else
\language=\csname l@#1\endcsname
\fi
#2}}
\providecommand{\BIBdecl}{\relax}
\BIBdecl

\bibitem{WHK2017}
M.~Wicker, X.~Huang, and M.~Kwiatkowska, ``Feature-guided black-box safety testing of deep neural networks,'' in \emph{International Conference on Tools and Algorithms for the Construction and Analysis of Systems}.\hskip 1em plus 0.5em minus 0.4em\relax Springer, 2018, pp. 408--426.

\bibitem{DLV}
X.~Huang, M.~Kwiatkowska, S.~Wang, and M.~Wu, ``Safety verification of deep neural networks,'' in \emph{29th International Conference on Computer Aided Verification}, ser. Lecture Notes in Computer Science, vol. 10426.\hskip 1em plus 0.5em minus 0.4em\relax Springer, 2017. doi: 10.1007/978-3-319-63387-9\_1 pp. 3--29.

\bibitem{Katz17}
G.~Katz, C.~W. Barrett, D.~L. Dill, K.~Julian, and M.~J. Kochenderfer, ``Reluplex: An efficient {SMT} solver for verifying deep neural networks,'' in \emph{29th International Conference on Computer Aided Verification}, ser. Lecture Notes in Computer Science, vol. 10426.\hskip 1em plus 0.5em minus 0.4em\relax Springer, 2017. doi: 10.1007/978-3-319-63387-9\_5 pp. 97--117.

\bibitem{VNNComp}
C.~Brix, M.~N. M{\"{u}}ller, S.~Bak, T.~T. Johnson, and C.~Liu, ``First three years of the international verification of neural networks competition {(VNN-COMP)},'' \emph{CoRR}, vol. abs/2301.05815, 2023. doi: 10.48550/arXiv.2301.05815

\bibitem{Falconmore22}
R.~Falconmore, ``On the role of explainability and uncertainty in ensuring safety of {AI} applications,'' Ph.D. dissertation, University of Oxford, {UK}, 2022.

\bibitem{Biggio13}
B.~Biggio, I.~Corona, D.~Maiorca, B.~Nelson, N.~Srndic, P.~Laskov, G.~Giacinto, and F.~Roli, ``Evasion attacks against machine learning at test time,'' in \emph{European Conference on Machine Learning and Knowledge Discovery in Databases}, ser. Lecture Notes in Computer Science, vol. 8190.\hskip 1em plus 0.5em minus 0.4em\relax Springer, 2013. doi: 10.1007/978-3-642-40994-3\_25 pp. 387--402.

\bibitem{Szegedy13}
\BIBentryALTinterwordspacing
C.~Szegedy, W.~Zaremba, I.~Sutskever, J.~Bruna, D.~Erhan, I.~J. Goodfellow, and R.~Fergus, ``Intriguing properties of neural networks,'' in \emph{2nd International Conference on Learning Representations}, 2014. [Online]. Available: \url{http://arxiv.org/abs/1312.6199}
\BIBentrySTDinterwordspacing

\bibitem{Biggio18}
B.~Biggio and F.~Roli, ``Wild patterns: Ten years after the rise of adversarial machine learning,'' in \emph{Proceedings of the 2018 {ACM} {SIGSAC} Conference on Computer and Communications Security}.\hskip 1em plus 0.5em minus 0.4em\relax {ACM}, 2018. doi: 10.1145/3243734.3264418 pp. 2154--2156.

\bibitem{Biggio12Poison}
B.~Biggio, B.~Nelson, and P.~Laskov, ``Poisoning attacks against support vector machines,'' in \emph{29th International Conference on Machine Learning}.\hskip 1em plus 0.5em minus 0.4em\relax icml.cc / Omnipress, 2012.

\bibitem{Wu2020}
M.~Wu, M.~Wicker, W.~Ruan, X.~Huang, and M.~Kwiatkowska, ``\BIBforeignlanguage{eng}{{A game-based approximate verification of deep neural networks with provable guarantees}},'' \emph{\BIBforeignlanguage{eng}{Theoretical Computer Science}}, vol. 807, pp. 298--329, 2020. doi: 10.1016/j.tcs.2019.05.046

\bibitem{Leino21globalrobust}
K.~Leino, Z.~Wang, and M.~Fredrikson, ``Globally-robust neural networks,'' in \emph{38th International Conference on Machine Learning}, ser. Proceedings of Machine Learning Research, vol. 139.\hskip 1em plus 0.5em minus 0.4em\relax {PMLR}, 2021, pp. 6212--6222.

\bibitem{Du20Interpretable}
M.~Du, N.~Liu, and X.~Hu, ``Techniques for interpretable machine learning,'' \emph{Commun. {ACM}}, vol.~63, no.~1, pp. 68--77, 2020. doi: 10.1145/3359786

\bibitem{molnar2020interpretable}
C.~Molnar, \emph{Interpretable machine learning}.\hskip 1em plus 0.5em minus 0.4em\relax Lulu. com, 2020.

\bibitem{bach2015pixel}
S.~Bach, A.~Binder, G.~Montavon, F.~Klauschen, K.-R. M{\"u}ller, and W.~Samek, ``On pixel-wise explanations for non-linear classifier decisions by layer-wise relevance propagation,'' \emph{PloS one}, vol.~10, no.~7, p. e0130140, 2015.

\bibitem{sundararajan2017axiomatic}
M.~Sundararajan, A.~Taly, and Q.~Yan, ``Axiomatic attribution for deep networks,'' in \emph{International conference on machine learning}.\hskip 1em plus 0.5em minus 0.4em\relax PMLR, 2017, pp. 3319--3328.

\bibitem{ribeiro2016should}
M.~T. Ribeiro, S.~Singh, and C.~Guestrin, ``"why should {I} trust you?": Explaining the predictions of any classifier,'' in \emph{22nd {ACM} {SIGKDD} International Conference on Knowledge Discovery and Data Mining}.\hskip 1em plus 0.5em minus 0.4em\relax {ACM}, 2016. doi: 10.1145/2939672.2939778 pp. 1135--1144.

\bibitem{ribeiro2018anchors}
------, ``Anchors: High-precision model-agnostic explanations,'' in \emph{Thirty-Second {AAAI} Conference on Artificial Intelligence}, vol.~32, no.~1, 2018.

\bibitem{IgnatievNM19}
A.~Ignatiev, N.~Narodytska, and J.~Marques{-}Silva, ``Abduction-based explanations for machine learning models,'' in \emph{Thirty-Third {AAAI} Conference on Artificial Intelligence}.\hskip 1em plus 0.5em minus 0.4em\relax {AAAI} Press, 2019. doi: 10.1609/aaai.v33i01.33011511 pp. 1511--1519.

\bibitem{LZM+21}
E.~L. Malfa, R.~Michelmore, A.~M. Zbrzezny, N.~Paoletti, and M.~Kwiatkowska, ``On guaranteed optimal robust explanations for {NLP} models,'' in \emph{Thirtieth International Joint Conference on Artificial Intelligence}.\hskip 1em plus 0.5em minus 0.4em\relax ijcai.org, 2021. doi: 10.24963/ijcai.2021/366 pp. 2658--2665.

\bibitem{Janusz23attribute}
A.~Janusz, D.~Slezak, S.~Stawicki, and K.~Stencel, ``A practical study of methods for deriving insightful attribute importance rankings using decision bireducts,'' \emph{Inf. Sci.}, vol. 645, p. 119354, 2023. doi: 10.1016/j.ins.2023.119354

\bibitem{wang2021statistically}
B.~Wang, S.~Webb, and T.~Rainforth, ``Statistically robust neural network classification,'' in \emph{Uncertainty in Artificial Intelligence}.\hskip 1em plus 0.5em minus 0.4em\relax PMLR, 2021, pp. 1735--1745.

\bibitem{Gehr18}
T.~Gehr, M.~Mirman, D.~Drachsler{-}Cohen, P.~Tsankov, S.~Chaudhuri, and M.~T. Vechev, ``{AI2:} safety and robustness certification of neural networks with abstract interpretation,'' in \emph{{IEEE} Symposium on Security and Privacy}.\hskip 1em plus 0.5em minus 0.4em\relax {IEEE} Computer Society, 2018. doi: 10.1109/SP.2018.00058 pp. 3--18.

\bibitem{Wang18}
S.~Wang, K.~Pei, J.~Whitehouse, J.~Yang, and S.~Jana, ``Efficient formal safety analysis of neural networks,'' in \emph{Annual Conference on Neural Information Processing Systems}, 2018, pp. 6369--6379.

\bibitem{Singh19}
G.~Singh, T.~Gehr, M.~P{\"{u}}schel, and M.~T. Vechev, ``An abstract domain for certifying neural networks,'' \emph{Proc. {ACM} Program. Lang.}, vol.~3, no. {POPL}, pp. 41:1--41:30, 2019. doi: 10.1145/3290354

\bibitem{Wang21beta}
S.~Wang, H.~Zhang, K.~Xu, X.~Lin, S.~Jana, C.~Hsieh, and J.~Z. Kolter, ``Beta-crown: Efficient bound propagation with per-neuron split constraints for neural network robustness verification,'' in \emph{Annual Conference on Neural Information Processing Systems}, 2021, pp. 29\,909--29\,921.

\bibitem{Matoba20Exact}
K.~Matoba and F.~Fleuret, ``Exact preimages of neural network aircraft collision avoidance systems,'' in \emph{Proceedings of the Machine Learning for Engineering Modeling, Simulation, and Design Workshop at Neural Information Processing Systems}, 2020, pp. 1--9.

\bibitem{Dathathri19Inverse}
S.~Dathathri, S.~Gao, and R.~M. Murray, ``Inverse abstraction of neural networks using symbolic interpolation,'' in \emph{Thirty-Third {AAAI} Conference on Artificial Intelligence}.\hskip 1em plus 0.5em minus 0.4em\relax {AAAI} Press, 2019. doi: 10.1609/aaai.v33i01.33013437 pp. 3437--3444.

\bibitem{Kotha23BoundPreimage}
S.~Kotha, C.~Brix, Z.~Kolter, K.~Dvijotham, and H.~Zhang, ``Provably bounding neural network preimages,'' \emph{CoRR}, vol. abs/2302.01404, 2023. doi: 10.48550/arXiv.2302.01404

\bibitem{ZWK23}
X.~Zhang, B.~Wang, and M.~Kwiatkowska, ``On preimage approximation for neural networks,'' \emph{CoRR}, vol. abs/2305.03686, 2023. doi: 10.48550/arXiv.2305.03686

\bibitem{Janusz23Diagnosis}
A.~Janusz, A.~Zalewska, L.~Wawrowski, P.~Biczyk, J.~Ludziejewski, M.~Sikora, and D.~Slezak, ``Brightbox - {A} rough set based technology for diagnosing mistakes of machine learning models,'' \emph{Appl. Soft Comput.}, vol. 141, p. 110285, 2023. doi: 10.1016/j.asoc.2023.110285

\bibitem{WongK18}
E.~Wong and J.~Z. Kolter, ``Provable defenses against adversarial examples via the convex outer adversarial polytope,'' in \emph{35th International Conference on Machine Learning}, ser. Proceedings of Machine Learning Research, vol.~80.\hskip 1em plus 0.5em minus 0.4em\relax {PMLR}, 2018, pp. 5283--5292.

\bibitem{CousotC92}
P.~Cousot and R.~Cousot, ``Abstract interpretation frameworks,'' \emph{J. Log. Comput.}, vol.~2, no.~4, pp. 511--547, 1992. doi: 10.1093/logcom/2.4.511

\bibitem{Wu2019}
M.~Wu and M.~Kwiatkowska, ``Robustness guarantees for deep neural networks on videos,'' in \emph{{IEEE/CVF} Conference on Computer Vision and Pattern Recognition}.\hskip 1em plus 0.5em minus 0.4em\relax Computer Vision Foundation / {IEEE}, 2020. doi: 10.1109/CVPR42600.2020.00039 pp. 308--317.

\bibitem{zhang2018crown}
H.~Zhang, T.~Weng, P.~Chen, C.~Hsieh, and L.~Daniel, ``Efficient neural network robustness certification with general activation functions,'' in \emph{Annual Conference on Neural Information Processing Systems}, 2018, pp. 4944--4953.

\bibitem{xu2021fast}
K.~Xu, H.~Zhang, S.~Wang, Y.~Wang, S.~Jana, X.~Lin, and C.~Hsieh, ``Fast and complete: Enabling complete neural network verification with rapid and massively parallel incomplete verifiers,'' in \emph{9th International Conference on Learning Representations}.\hskip 1em plus 0.5em minus 0.4em\relax OpenReview.net, 2021.

\bibitem{CousotC77}
P.~Cousot and R.~Cousot, ``Abstract interpretation: {A} unified lattice model for static analysis of programs by construction or approximation of fixpoints,'' in \emph{Fourth {ACM} Symposium on Principles of Programming Languages}.\hskip 1em plus 0.5em minus 0.4em\relax {ACM}, 1977. doi: 10.1145/512950.512973 pp. 238--252.

\bibitem{MirmanGV18}
M.~Mirman, T.~Gehr, and M.~T. Vechev, ``Differentiable abstract interpretation for provably robust neural networks,'' in \emph{35th International Conference on Machine Learning}, ser. Proceedings of Machine Learning Research, vol.~80.\hskip 1em plus 0.5em minus 0.4em\relax {PMLR}, 2018, pp. 3575--3583.

\bibitem{SinghGMPV18}
G.~Singh, T.~Gehr, M.~Mirman, M.~P{\"{u}}schel, and M.~T. Vechev, ``Fast and effective robustness certification,'' in \emph{Annual Conference on Neural Information Processing Systems}, 2018, pp. 10\,825--10\,836.

\bibitem{Albarghouthi21}
A.~Albarghouthi, ``Introduction to neural network verification,'' \emph{Found. Trends Program. Lang.}, vol.~7, no. 1-2, pp. 1--157, 2021. doi: 10.1561/2500000051

\bibitem{ehlers2017formal}
R.~Ehlers, ``Formal verification of piece-wise linear feed-forward neural networks,'' in \emph{Automated Technology for Verification and Analysis}.\hskip 1em plus 0.5em minus 0.4em\relax Springer International Publishing, 2017, pp. 269--286.

\bibitem{DuttaJST18}
S.~Dutta, S.~Jha, S.~Sankaranarayanan, and A.~Tiwari, ``Output range analysis for deep feedforward neural networks,'' in \emph{10th International Symposium on {NASA} Formal Methods}, ser. Lecture Notes in Computer Science, vol. 10811.\hskip 1em plus 0.5em minus 0.4em\relax Springer, 2018. doi: 10.1007/978-3-319-77935-5\_9 pp. 121--138.

\bibitem{FischettiJ18}
M.~Fischetti and J.~Jo, ``Deep neural networks and mixed integer linear optimization,'' \emph{Constraints An Int. J.}, vol.~23, no.~3, pp. 296--309, 2018. doi: 10.1007/s10601-018-9285-6

\bibitem{Tjeng19}
V.~Tjeng, K.~Y. Xiao, and R.~Tedrake, ``Evaluating robustness of neural networks with mixed integer programming,'' in \emph{7th International Conference on Learning Representations}.\hskip 1em plus 0.5em minus 0.4em\relax OpenReview.net, 2019.

\bibitem{Vielma15}
J.~P. Vielma, ``Mixed integer linear programming formulation techniques,'' \emph{{SIAM} Rev.}, vol.~57, no.~1, pp. 3--57, 2015. doi: 10.1137/130915303

\bibitem{BunelLTTKK20}
R.~Bunel, J.~Lu, I.~Turkaslan, P.~H.~S. Torr, P.~Kohli, and M.~P. Kumar, ``Branch and bound for piecewise linear neural network verification,'' \emph{J. Mach. Learn. Res.}, vol.~21, pp. 42:1--42:39, 2020.

\bibitem{WangPWYJ18}
S.~Wang, K.~Pei, J.~Whitehouse, J.~Yang, and S.~Jana, ``Formal security analysis of neural networks using symbolic intervals,'' in \emph{27th {USENIX} Security Symposium}.\hskip 1em plus 0.5em minus 0.4em\relax {USENIX} Association, 2018, pp. 1599--1614.

\bibitem{SotoudehTT23}
M.~Sotoudeh, Z.~Tao, and A.~V. Thakur, ``Syrenn: {A} tool for analyzing deep neural networks,'' \emph{Int. J. Softw. Tools Technol. Transf.}, vol.~25, no.~2, pp. 145--165, 2023. doi: 10.1007/s10009-023-00695-1

\bibitem{Albarghouthi13interpolants}
A.~Albarghouthi and K.~L. McMillan, ``Beautiful interpolants,'' in \emph{25th International Conference on Computer Aided Verification}, ser. Lecture Notes in Computer Science, vol. 8044.\hskip 1em plus 0.5em minus 0.4em\relax Springer, 2013. doi: 10.1007/978-3-642-39799-8\_22 pp. 313--329.

\bibitem{ruan2019hamming}
W.~Ruan, M.~Wu, Y.~Sun, X.~Huang, D.~Kroening, and M.~Kwiatkowska, ``Global robustness evaluation of deep neural networks with provable guarantees for the hamming distance,'' in \emph{Twenty-Eighth International Joint Conference on Artificial Intelligence}.\hskip 1em plus 0.5em minus 0.4em\relax International Joint Conferences on Artificial Intelligence Organization, 2019, pp. 5944--5952.

\bibitem{baluta2021quantitative}
T.~Baluta, Z.~L. Chua, K.~S. Meel, and P.~Saxena, ``Scalable quantitative verification for deep neural networks,'' in \emph{43rd {IEEE/ACM} International Conference on Software Engineering}.\hskip 1em plus 0.5em minus 0.4em\relax {IEEE}, 2021. doi: 10.1109/ICSE43902.2021.00039 pp. 312--323.

\bibitem{yang2021quantitative}
P.~Yang, R.~Li, J.~Li, C.~Huang, J.~Wang, J.~Sun, B.~Xue, and L.~Zhang, ``Improving neural network verification through spurious region guided refinement,'' in \emph{27th International Conference on Tools and Algorithms for the Construction and Analysis of Systems}, ser. Lecture Notes in Computer Science, vol. 12651.\hskip 1em plus 0.5em minus 0.4em\relax Springer, 2021, pp. 389--408.

\bibitem{la2020assessing}
E.~L. Malfa, M.~Wu, L.~Laurenti, B.~Wang, A.~Hartshorn, and M.~Kwiatkowska, ``Assessing robustness of text classification through maximal safe radius computation,'' in \emph{Findings of the Association for Computational Linguistics}, ser. Findings of {ACL}, vol. {EMNLP} 2020.\hskip 1em plus 0.5em minus 0.4em\relax Association for Computational Linguistics, 2020. doi: 10.18653/v1/2020.findings-emnlp.266 pp. 2949--2968.

\bibitem{LK22}
E.~L. Malfa and M.~Kwiatkowska, ``The king is naked: On the notion of robustness for natural language processing,'' in \emph{Thirty-Sixth {AAAI} Conference on Artificial Intelligence}.\hskip 1em plus 0.5em minus 0.4em\relax {AAAI} Press, 2022, pp. 11\,047--11\,057.

\bibitem{BPW+22}
E.~Benussi, A.~Patan{\`{e}}, M.~Wicker, L.~Laurenti, and M.~Kwiatkowska, ``Individual fairness guarantees for neural networks,'' in \emph{Thirty-First International Joint Conference on Artificial Intelligence}.\hskip 1em plus 0.5em minus 0.4em\relax ijcai.org, 2022. doi: 10.24963/ijcai.2022/92 pp. 651--658.

\bibitem{michelmore2019uncertainty}
R.~Michelmore, M.~Wicker, L.~Laurenti, L.~Cardelli, Y.~Gal, and M.~Kwiatkowska, ``Uncertainty quantification with statistical guarantees in end-to-end autonomous driving control,'' in \emph{{IEEE} International Conference on Robotics and Automation}.\hskip 1em plus 0.5em minus 0.4em\relax {IEEE}, 2020. doi: 10.1109/ICRA40945.2020.9196844 pp. 7344--7350.

\bibitem{wicker20}
M.~Wicker, L.~Laurenti, A.~Patane, and M.~Kwiatkowska, ``Probabilistic safety for bayesian neural networks,'' in \emph{Thirty-Sixth Conference on Uncertainty in Artificial Intelligence}, ser. Proceedings of Machine Learning Research, vol. 124.\hskip 1em plus 0.5em minus 0.4em\relax {AUAI} Press, 2020, pp. 1198--1207.

\bibitem{wicker2021certification}
M.~Wicker, L.~Laurenti, A.~Patane, N.~Paoletti, A.~Abate, and M.~Kwiatkowska, ``Certification of iterative predictions in bayesian neural networks,'' in \emph{Thirty-Seventh Conference on Uncertainty in Artificial Intelligence}, ser. Proceedings of Machine Learning Research, vol. 161.\hskip 1em plus 0.5em minus 0.4em\relax {AUAI} Press, 2021, pp. 1713--1723.

\bibitem{gourdeau2019hardness}
P.~Gourdeau, V.~Kanade, M.~Kwiatkowska, and J.~Worrell, ``On the hardness of robust classification,'' in \emph{Advances in Neural Information Processing Systems}, 2019, pp. 7444--7453.

\bibitem{gourdeau2021hardness}
------, ``On the hardness of robust classification,'' \emph{Journal of Machine Learning Research}, vol.~22, 2021.

\bibitem{gourdeau2022sample}
------, ``Sample complexity bounds for robustly learning decision lists against evasion attacks,'' in \emph{Thirty-First International Joint Conference on Artificial Intelligence}.\hskip 1em plus 0.5em minus 0.4em\relax ijcai.org, 2022. doi: 10.24963/ijcai.2022/419 pp. 3022--3028.

\bibitem{gourdeau2022when}
------, ``When are local queries useful?'' in \emph{Advances in Neural Information Processing Systems}, 2022.

\end{thebibliography}

\end{document}